\documentclass[preprint,12pt]{elsarticle}

\usepackage{amssymb}
\usepackage[hidelinks]{hyperref}
\usepackage{aliases}

\journal{}

\begin{document}

\begin{frontmatter}



\title{Solution of Physics-based Bayesian Inverse Problems with Deep Generative Priors}


\author[inst1, inst2]{Dhruv V. Patel}
\ead{dhruvvpa@usc.edu}
\affiliation[inst1]{organization={Aerospace and Mechanical Engineering, University of Southern California}, 
            city={Los Angeles},
            postcode={90089}, 
            state={CA},
            country={USA}}
            
\affiliation[inst2]{organization={Department of Mechanical Engineering, Stanford University}, 
            city={Stanford},
            postcode={94305}, 
            state={CA},
            country={USA}}

\author[inst1]{Deep Ray}
\author[inst1]{Assad A. Oberai}


\begin{abstract}
Inverse problems are ubiquitous in nature, arising in almost all areas of science and engineering ranging from geophysics and climate science to astrophysics and biomechanics. One of the central challenges in solving inverse problems is tackling their ill-posed nature. Bayesian inference provides a principled approach for overcoming this by formulating the inverse problem into a statistical framework. However, it is challenging to apply when inferring fields that have discrete representations of large dimensions (the so-called ``curse of dimensionality'') and/or when prior information is available only in the form of previously acquired solutions.  In this work, we present a novel method for efficient and accurate Bayesian inversion using deep generative models. Specifically, we demonstrate how using the approximate distribution learned by a Generative Adversarial Network (GAN) as a prior in a Bayesian update and reformulating the resulting inference problem in the low-dimensional latent space of the GAN, enables the efficient solution of large-scale Bayesian inverse problems. Our statistical framework preserves the underlying physics and is demonstrated to yield accurate results with reliable uncertainty estimates, even in the absence of information about underlying noise model, which is a significant challenge with many existing methods. We demonstrate the effectiveness of proposed method on a variety of inverse problems which include both synthetic as well as experimentally observed data.

\end{abstract}


\begin{highlights}
\item Efficient algorithm leveraging ML and physics to solve statistical inverse problems.
\item	Utilizes GANs to construct sample-based priors and achieve dimensionality reduction.
\item Suitable for high dimensional Bayesian inversion involving complex prior density.
\item Effective for a range of inverse problems involving experimental and in-silico data.


\end{highlights}

\begin{keyword}
Bayesian inference \sep Inverse problems \sep Uncertainty Quantification (UQ) \sep Model order reduction \sep Markov Chain Monte Carlo (MCMC) \sep Elastography
\PACS 0000 \sep 1111
\MSC 0000 \sep 1111
\end{keyword}

\end{frontmatter}


\section{Introduction}
\label{sec:intro}

The definition of what constitutes an inverse problem is not very precise. In his article titled \textit{Inverse Problems}, Joseph Keller \cite{keller1976} defines two problems as inverses of each other if the formulation of one involves the complete or partial solution of the other. Under these conditions the problem that has been studied extensively is termed as the ``direct problem'', and the one that is newer, and not as well studied, is called the ``inverse problem.'' The author then gives some serious and some not-so-serious examples of inverse problems. 
A typical direct problem in engineering is one of finding the temperature field in a Fourier solid given the knowledge of the heat source, boundary conditions, and thermal conductivity. The corresponding inverse problem is that of finding the thermal conductivity field given the temperature field. Similar inverse problems are at the core of techniques that infer mechanical properties of tissue in elastography \cite{doyley2014elastography,barbone2010review}, X-ray tomography, inverse acoustic/electromagnetic scattering \cite{kirsch2021introduction,isakov2006inverse}, waveform inversion in geophysics \cite{Snieder1999inverse, Gouveia1997}, data assimilation in climate modeling \cite{Jackson2004, Huang2005inverse}, and remote sensing in astronomy \cite{Craig1986InversePI, AsensioRamos2007}. 

In addition to being newer and less well-studied, most inverse problems are ill-posed in the sense of Hadamard \cite{hadamard1902problemes}. That is, they either have multiple solutions (non-unique), no solutions (non-existence),
or have solutions that vary significantly in response to small perturbations in measured data (unstable). Further, they are driven by measured input that is almost always corrupted with stochastic noise. 
One approach to address these challenges relies on ``regularizing'' the inverse problem \cite{engl1996regularization, peyre2008non}, which amounts to solving an alternate well-posed problem whose solution is close to that of the original inverse problem. Another approach, which is often referred to as Bayesian inference, involves viewing the inverse problem as a stochastic inference problem and using Bayes' rule to update the prior probability distribution of the solution with the knowledge gained from the measurement \cite{kaipio2006statistical,dashti2016bayesian}. A significant advantage of Bayesian inference is that it provides quantitative estimates of the probability distribution of the solution which can be used to quantify the confidence in the solution. This could be crucial for applications where high-stake decisions are made based on the output of the inverse problem.

The application of Bayesian inference to practical problems leads to two significant challenges. 
The first involves selecting an appropriate prior probability distribution that encodes all the available knowledge about the solution. Since Bayesian inference requires an explicit formula of the prior distribution, most practitioners use simple distributions as the prior density. These include the multivariate Gaussian distribution, Laplace or Gaussian processes with specified covariance kernels, and hierarchical priors. This ensures that the resulting posterior can be probed in a tractable manner using either analytical (conjugate prior) or numerical (variational inference or Monte Carlo) methods. However, the true prior distribution may be much more complex (see Figure \ref{fig:priors} for example) and using simple handcrafted prior density for such applications may lead to biased posterior distribution. Furthermore, often prior knowledge about the field to be inferred is available implicitly in terms of historical data or previously acquired solutions. Translating this implicit \textit{qualitative domain knowledge} into an explicit \textit{quantitative formula} that can be used as prior density is challenging. 

The second challenge in Bayesian inference is the so-called \textit{curse of dimensionality}, which occurs when the dimension of the inferred solution vector is large. In this case, sampling from the posterior distribution using methods like Markov Chain Monte Carlo (MCMC) becomes prohibitively expensive. This also leads to an inherent algorithmic challenge -- designing an MCMC algorithm that scales well to high dimensions is an extremely difficult task \cite{Betancourt2017ACI}. 


In this paper we tackle these two challenges associated with Bayesian inference. We consider problems in which prior knowledge about the solution is available in the form of many independent and identically distributed samples drawn from the true prior distribution. Such samples might be available in the form of historical data collected from a series of experiments. For example, the measurement of permeability field in petroleum reservoirs, CT scans of material microstructures, and MR images of the human brain. We describe how a generative adversarial network (GAN) \cite{goodfellow2014generative}, a type of a deep generative model, may be used to encode this information to perform efficient Bayesian inference. This is accomplished by mapping a simple distribution in the low-dimensional latent space of the GAN to the complex prior distribution in the space of the solution vector. Further, we map back the expression of the posterior distribution into the latent space of the GAN and utilize the fact that its dimension is typically much smaller than that of the solution space to achieve dimension reduction. In addition to reducing the effective dimension of the posterior distribution, this reformulation allows for effective exploration of posterior with simpler MCMC algorithms, because it is empirically observed that the posterior distribution in the latent space has a much simpler geometrical structure (unimodal) than in the original space.
We demonstrate the broad applicability of these ideas by solving inverse problems with different direct problems (heat equation, Navier's equations for elasticity and the Radon transform), different prior distributions (simple geometrical features, MNIST digits, material micro-structure and the Shepp-Logan phantom), and synthetic and \green{experimentally} measured data. When possible, we compare our results with ``true'' values, other commonly used \green{inference} methods, and experimental measurements. 

In recent years deep generative models have shown considerable promise in learning complex probability distributions and have been used to solve deterministic and stochastic inverse problems arising in physical sciences, computer vision, and medical imaging domains \cite{Lindgren2020, goh2019solving, rizzuti2020parameterizing, bora2017compressed, ongie2020deep}. These works use different deep generative models based on variational inference \cite{kingma2013auto, rezende2015variational}, normalizing flows \cite{Dinh2016}, and adversarial learning \cite{goodfellow2014generative}. In \cite{adler2018deep}, GANs are used to directly learn the posterior distribution of the inverse problem. This is accomplished by constructing a generator whose output is conditioned on the measurement vector. In \cite{yang2019adversarial}, also a conditional version of a GAN is used; however the output of the generator is conditioned on the spatial coordinates for the problem, and the direct problem is also solved within the network by appending to the loss function a term that penalizes the residual of direct problem. Physics informed GANs were designed in \cite{Yang2020pigan} to solve forward, inverse, and mixed stochastic problems, with the underlying physical law encoded into the GAN training via a penalty term in the loss function. An overview of different uncertainty quantification methods used in scientific machine learning is provided in \cite{psaros2022uncertainty}. In contrast to these methods, we explore the use of GANs in approximating the prior in Bayesian inference. This is an extension of our recent work \cite{patel2021gan} to problems where the direct map, which is derived from the underlying physics, is known. We also note that in Bayesian inference the fields of sample based priors \cite{Vauhkonen1997, Calvetti2005}, and dimension-reduction \cite{Li2006efficient, Marzouk2009, Lieberman2010} have a rich history of methods that do not employ the more recent machine learning algorithms.

\paragraph{Our contribution} In this manuscript we focus on the problem of solving a physics-based Bayesian inverse problem, and make the following contributions:

\begin{itemize}
    \item We propose a novel way of characterizing the prior distribution in Bayesian inference when the prior information is available in the form of previously acquired solutions or historical data.
    \item We reformulate the resulting posterior distribution in the low-dimensional latent space of the GAN for efficient posterior exploration, thereby enabling large-scale Bayesian inversion. This reformulation is informed by physics (for the likelihood contribution) and is driven by data (for the prior contribution).
    
    \item We demonstrate the effectiveness of the proposed algorithm on a variety of high dimensional stochastic inverse problems involving synthetic and experimental data.
\end{itemize}

The layout of the remainder of the manuscript is as follows. In the following section we define the problem of interest and develop the new method of performing Bayesian inversion using GAN-based priors in an unsupervised fashion. Thereafter, in Section \ref{sec:results}, we apply this method to a series of stochastic inverse problems: inverse heat conduction (thermal conductivity and source inversion), inverse Radon transform, and elasticity imaging, and compare its performance against the ground truth and other methods. We end conclusions in Section \ref{sec:conclusions}.




\section{Mathematical Formulation}

\subsection{\green{Bayesian inference}}\label{sec:bayes_formulation}

Consider the following direct model
\begin{equation}\label{eqn:forward_map}
	\fmap : \infv \mapsto \meas, \quad \infv \in \dinfv \subset \Ro^{\Ninfv}, \quad \meas \in \dmeas \subset \Ro^{\Nmeas},
\end{equation}
where $\meas$ is the measured response for some input $\infv$. 
Direct problems are generally well studied and have robust numerical solvers. On the other hand, inverse problems, where we wish to infer $\infv$ given a measurement $\meas$, are challenging. Bayesian inference provides a way of solving inverse problems with quantified uncertainty estimates. Within this approach the inferred field and observations are modeled as realizations of the random variables $\infvRV$ and $\measRV$, respectively. It is often assumed that the measurement is corrupted by an additive noise, i.e., $\nmeas = \meas + \noise$, where $\noise$ is the noise and $\pn$ is its distribution. A prior distribution $\priorinfv(\infv)$, which encodes the knowledge about $\infvRV$ \textit{prior} to observing $\nmeas$, is assumed. This is used along with the likelihood of observing the measurement $\nmeas$, given $\infvRV=\infv$, that is $\plike(\nmeas | \infv)$. Using Bayes' rule, this yields the following expression for the posterior distribution of $\infvRV$,
\begin{equation}\label{eqn:bayes}
    \postinfv(\infv |\nmeas ) = \frac{\plike(\nmeas|\infv)\priorinfv(\infv)}{\pmeas(\nmeas)}  \propto \pn(\nmeas - \fmap(\infv)) \priorinfv(\infv),
\end{equation}
where $\pmeas(\nmeas)$ is called the evidence. The likelihood term, $\plike(\nmeas|\infv) = \pn(\nmeas - \fmap(\infv))$, injects physics into the framework via the direct operator $\fmap$. 
Statistical inference involves characterizing the posterior distribution by computing the expectation of a statistical quantity of interest (QoI) (defined by $l(\infv)$) with respect to this posterior,
\begin{equation}\label{eqn:qoi_post}
    \ex{\infv \sim \postinfv(\infv|\nmeas)}{l(\infv)} = \int_{\dinfv} l(\infv) \postinfv(\infv|\nmeas) d \infv.
\end{equation}
For most problems of practical interest, the dimension of the inferred field is typically large $(\bO(10^3-10^6))$,  and the integral \eqref{eqn:qoi_post} cannot be computed using a standard numerical/quadrature-based method; instead it is approximated using MCMC methods \cite{Brooks2011}.



\begin{figure}[!htbp]
\centering
\includegraphics[width=\textwidth]{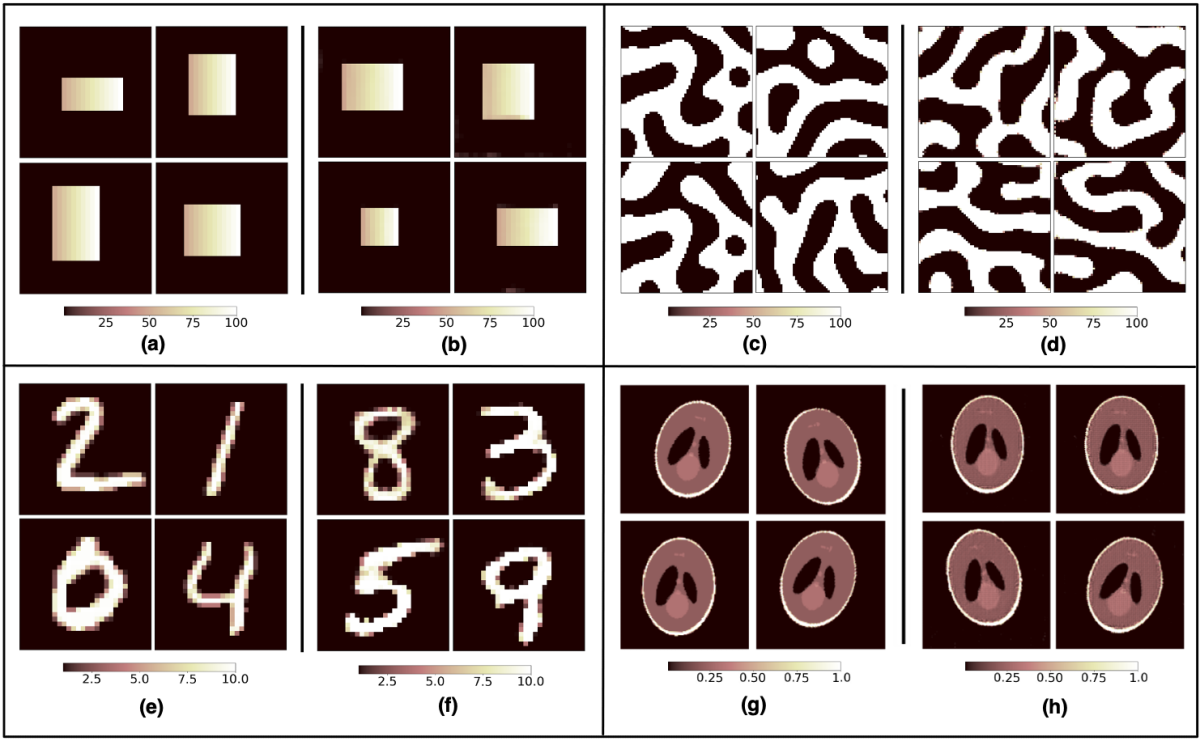}
\caption{Typical samples from different prior distributions: (a) Parametric rectangular thermal conductivity, (c) Bi-phase material microstructure, (e) MNIST handwritten digits, (g) Shepp-Logan phantoms. The samples in (b), (d), (f) and (h) are counterparts to (a), (c), (e) and (g) respectively, generated by a WGAN-GP.}
\label{fig:priors}
\end{figure}

There are two major challenges in using the Bayesian framework described above. First, constructing a closed-form expression of the prior, $\priorinfv(\infv)$, is challenging, especially when the prior information is available through samples. Consider, for example, samples of the microstructure of a material shown in Figure \ref{fig:priors}(c) or the Shepp-Logan phantom of the human head shown in Figure \ref{fig:priors}(g). It should be clear that the variability in this data cannot be modeled  using standard distributions, or their mixtures. Second, for problems with high dimensional parameter space, the number of MCMC samples required for a well-mixed Markov chain is very large, making it computationally expensive to accurately approximate any desired quantity of interest (QoI).
In the development below, we describe how a GAN, a type of a deep learning algorithm, can be used to address these challenges. 

\begin{figure}[!htbp]
\centering
\includegraphics[width=\textwidth]{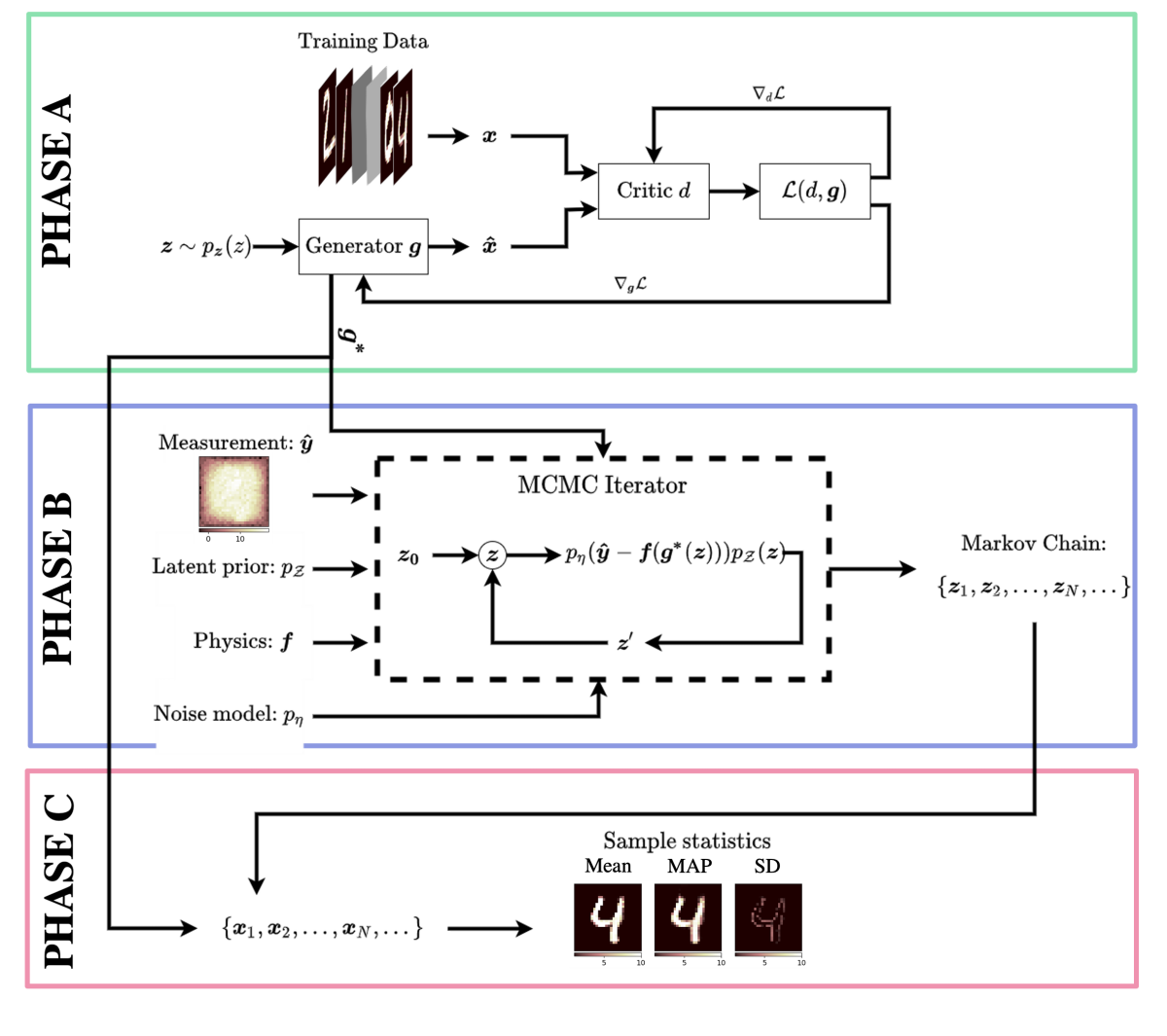}
\caption{Schematic diagram of the inference algorithm. PHASE A: A WGAN-GP is trained using previously acquired samples of the field that is to be inferred. PHASE B: This generator, the measurement, the direct model, and the noise distribution are used in an MCMC iterator which learns the posterior distribution in latent space. PHASE C: The converged Markov chain is used to sample from the posterior distribution in the latent space and these samples are passed through the trained generator to yield samples of the inferred field.}
\label{fig:schematic}
\end{figure}

\subsection{Generative Adversarial Networks}
\label{sec:BayInf_GAN}

GANs typically comprise two neural networks (generator and discriminator) that are trained adversarialy to learn a target distribution $\pinfv(\infv)$ and generate samples from it. The generator $\gen(\lsv; \btheta)$ maps a latent variable $\lsv \in \dlsv \subset \Ro^{\Nlsv}$ sampled from $\plsv$ to the space of the target field $\infv \in \dinfv \subset \Ro^{\Ninfv} $, where $\btheta \in \Ro^{\Ngen}$ denote the trainable parameters. Typically, $\plsv$ is selected from an easy-to-sample-from distribution (such as multivariate Gaussian or uniform),  with $\Nlsv \ll \Ninfv $. The second scalar-valued neural network is known as the discriminator $\disc(\infv; \bphi)$, with parameters $\bphi \in \Ro^{\Ndisc}$, and is used to measure the discrepancy between the  distribution $\pinfv(\infv)$ and the distribution learned by the generator. 


A number of GAN models have been developed over the years. They differ from each other on the basis of their loss function (see \cite{goodfellow2014generative, poole2016improved, nowozin2016fgan} for example).  In order to use a GAN as a prior in Bayesian inference (the main objectives of this work), the convergence of the density induced by the generator of the GAN to the true prior density $\priorinfv$ is required. The Wasserstein GAN (WGAN) \cite{arjovsky2017wasserstein_proc}, which minimizes the Wasserstein-1 distance between these two densities, satisfies this key property and is therefore used in this work. Specifically, we use the improved version of the WGAN with a Gradient Penalty (WGAN-GP) \cite{gulrajani2017improved}, whose loss function is given by
\begin{equation}\label{eqn:wganlossgp}
\begin{aligned}
\loss(\disc,\gen) &=  \ex{\infv \sim \priorinfv (\infv)}{\disc(\infv)} - \ex{\lsv \sim \plsv(\lsv)}{\disc(\gen(\lsv))} + \underbrace{ \lambda \ex{\hat{\infv} \sim \hat{p}_X(\infv)}{(\|\nabla_{\hat{\infv}} \disc(\hat{\infv})\|_2 - 1)^2}}_{\text{gradient penalty}}.
\end{aligned}
\end{equation}
In \eqref{eqn:wganlossgp}, the first term is the expectation of the discriminator over the set of prior samples, the second term is the expectation over the set of fake samples, $\lambda$ is the penalty parameter and $\hat{p}_X(\infv)$ defines a uniform sampling along straight lines between pairs of points sampled from $\priorinfv(\infv)$ and the push-forward of $\plsv(\lsv)$ by $\gen$. The gradient penalty constrains the discriminator to be a 1-Lipschitz function. The WGAN-GP is trained by solving the following min-max problem, 
\begin{equation}\label{eqn:minimax}
(\disc^*, \gen^*) = \argmin{\gen } \; \argmax{\disc } \; \loss(\disc,\gen).
\end{equation}
As described in \cite{arjovsky2017wasserstein_proc,gulrajani2017improved}, the inner maximization problem under the Lipschitz constraint on the discriminator leads to an approximation of the Wasserstein-1 distance between $\pinfv$ and the push-forward of $\plsv$ by a given generator. The outer minimization problem corresponds to finding a generator that minimizes this Wasserstein-1 distance. In our work, we use the WGAN-GP such that the push-forward of $\plsv$ approximates the prior distribution.

\subsection{Bayesian inference with GAN priors}\label{sec:GAN_priors}
Since convergence of distributions in the Wasserstein-1 metric is equivalent to weak convergence (Theorem 6.9 in \cite{villani2008optimal}), for a perfectly trained generator, $\gen^*$, trained using samples from the prior distribution, $\priorinfv$,  we have,
\begin{equation}\label{eqn:mom_match1}
\ex{\infv \sim \priorinfv(\infv)}{m(\infv)} = 
\ex{\lsv \sim \plsv(\lsv)}{m(\gen^*(\lsv))} \quad \forall \ m \in C_b(\dinfv),
\end{equation}
where $C_b(\cdot)$ is the space of continuous bounded functions. 

Equation \eqref{eqn:mom_match1} can be converted into an expression for the posterior distribution, assuming that $\pn$ is continuous and bounded. This is accomplished by setting $m(\cdot) = l(\cdot) \pn(\nmeas - \fmap(.))/\pmeas(\nmeas)$ in (\ref{eqn:mom_match1})
where $l \in C_b(\dinfv)$, and using \eqref{eqn:bayes}, to arrive at
\begin{equation}\label{eqn:post_dist_match}
\ex{\infv \sim \postinfv(\infv)}{l(\infv)} =  \ex{\lsv \sim \postlsv(\lsv)}{l(\gen^*(\lsv))}.            		 
\end{equation}
Here $\postlsv$ is the posterior distribution of $\lsvRV$ given $\nmeas$ 
and is defined as,
\begin{equation}\label{eqn:bayes_meas_lsv}
    \postlsv(\lsv | \nmeas) = \frac{\pn(\nmeas - \fmap(\gen^*(\lsv))) \plsv(\lsv)}{\pmeas(\nmeas)}.
\end{equation}
Expression \eqref{eqn:post_dist_match} implies that statistics with respect to $\postinfv(\infv)$ can be evaluated by sampling $\lsv$ from $\postlsv$ and passing the samples through $\gen^*$. Since the expression for $\plsv(\lsv)$ is known, and $\Nlsv \ll \Ninfv$, we can efficiently sample from $\postlsv(\lsv)$ using an MCMC algorithm.

As shown in Figure \ref{fig:schematic}, the inference algorithm described above involves three phases. In Phase A, the WGAN-GP is trained using previously acquired samples of the field to be inferred. In Phase B, the trained generator, the measured field, the direct map, and the noise distribution are used to generate a Markov chain which samples from an approximation to the posterior distribution in the latent space. In Phase C, this Markov chain and the trained generator are used to generate samples to evaluate the desired QoIs as described below.

\subsection{Expression for QoIs and the MAP estimate}\label{sec:QOI}
In this section we describe how Equation (\ref{eqn:post_dist_match}) can be utilized to compute some specific QoIs and the maximum a-posteriori (MAP) estimate. 

\paragraph{Mean} An estimate of the mean of the posterior distribution, $ \postinfv(\infv|\nmeas)$, can be computed by using $l(\bm{x}) = \bm{x}$ in Equation \eqref{eqn:post_dist_match}. That is
\begin{equation}\label{mean_estimate}
    \overline{\infv} = \ex{\bm{\infv \sim \postinfv}}{\infv} = \ex{\bm{\lsv \sim \postlsv}}{\bm{g}^*(\lsv)}.
\end{equation}

\paragraph{Standard deviation} In order to compute the standard deviation estimate with respect to the posterior distribution, we first compute the second moment,
\begin{equation}
    \overline{\bm{m}} = \ex{\bm{\infv \sim \postinfv}}{\infv^2} = \ex{\bm{\lsv \sim \postlsv}}{(\bm{g}^*(\lsv))^2},
\end{equation}
where the square of a vector is understood to be the square of each of its components. Thereafter, we use the mean and the second moment to evaluate the component-wise standard deviation,
\begin{equation}
    \bm{\sigma} = \sqrt{\overline{\bm{m}} - \overline{\infv}^2}.
\end{equation}

\paragraph{MAP Estimate} 
In the case, where we are only interested in finding the most probable answer, and not necessarily uncertainty, directly finding the MAP estimate could be useful. Since the explicit expression for the prior $\priorinfv(\infv)$ is unknown, it is not possible to use Equation \eqref{eqn:bayes} to determine the value of the posterior distribution (even up to a multiplicative constant). Therefore, we cannot directly evaluate the MAP estimate, $\infv^{\rm MAP}$, within our approach. However, it is possible to estimate $\lsv^{\rm MAP}$ using the expression for $\postlsv(\lsv|\nmeas)$ (Equation (\ref{eqn:pz_post})). For every sample generated by the MCMC chain, we evaluate the log of this distribution (up to an additive constant), and of those, designate the sample with the largest value as $\lsv^{\rm MAP}$. Thereafter we treat $\bm{g}^*(\lsv^{\rm MAP})$ as an approximation to $\infv^{\rm MAP}$. 


\begin{remark} In the special case of additive Gaussian noise ($\pn$) and Gaussian latent distribution ($\plsv$), $\lsv^{\rm MAP}$ can be determined using a simple optimization algorithm. 
We consider the case when the components of the latent vector are i.\,i.\,d.\,sampled from a normal distribution with zero mean and unit variance. This is often the case in many typical GAN applications. Further, we assume that the components of the noise vector are defined by a normal distribution with zero mean and a covariance matrix $\bm{\Sigma}$. Using these assumptions in (\ref{eqn:bayes_meas_lsv}), we have 
\begin{eqnarray}
\postlsv(\lsv|\nmeas) \propto \exp{\Big( -\overbrace{\frac{1}{2}\big( | \bm{\Sigma}^{-1/2} (\nmeas - \bm{f}(\bm{g}^*(\lsv)))|^2  + |\lsv|^2 \big)}^{\equiv r(\lsv)} \Big)}.
\label{eqn:pz_post}
\end{eqnarray}
The MAP estimate for this distribution is obtained by minimizing the negative of the argument of the exponential. That is 
\begin{eqnarray}
\lsv^{\rm MAP} = \amin_{\lsv} r(\lsv). \label{eqn:zmap}
\end{eqnarray}

This could be useful, for example, in scenarios where one is interested in solving only deterministic inverse problem.
\end{remark}


It is worth noting that in practice, the GANs are trained by minimizing an empirical estimator of the Wasserstein-1 distance.  This means we can only approximately satisfy the Equation \ref{eqn:mom_match1}. Theoretical results about the convergence of such estimators, in the broader context of integral probability metrics, have been explored in \cite{Sriperumbudur2009OnIP, Arora2018DoGL, Uppal2019NonparametricDE} and the rates of convergence for the Wasserstein estimator have also been derived. Preliminary statistical error estimates of the prior density learned by a WGAN with the true prior have been explored in \cite{patel2020bayesian}. Such results can be combined with available convergence results for MCMC algorithms to obtain error estimates in approximating the posterior statistics. However, this is beyond the scope of the present paper and will be considered in future works. Further, such results typically do not account for the errors due to the stochastic gradient based optimizer (such as Adam) used to train the network, which adds yet another layer of complexity in being able to use these theoretical results for practical purposes. In practice, we compute and monitor the empirical Wasserstein-1 distance and terminate the training once this distance has decreased to a reasonably low value and has saturated.

\section{Numerical validation and results}\label{sec:results}

We consider several physics-based applications to demonstrate the effectiveness of the proposed algorithm. These can be classified based on the following criteria,
\begin{enumerate}
    \item Direct Model: we consider the diffusion equation, the linear elasticity equation, and the Radon transform.
    \item  Likelihood distribution: we consider three different scenarios commonly encountered in practice while solving inference problems: 
    
    \begin{enumerate}
        \item The likelihood distribution (Gaussian $\mathcal{N}(\bm{0}, \sigma^2\bm{\mathds{1}})$) and its parameters ($\sigma$) are known. This corresponds to the inverse heat conduction problem of thermal conductivity inversion and source inversion (Section \ref{sec:heat_conduction} and \ref{sec:source_inversion}, respectively)
        
        \item The likelihood distribution ($\mathcal{N}(\cdot, \cdot)$) is known but its parameters ($\sigma$) are not. This corresponds to the inverse radon transform problem (Section \ref{sec:radon_transform}), where the knowledge of correct noise distribution is assumed, but an incorrect value of its parameter (variance) is used
        
        \item The likelihood distribution is not known and must be guessed, which corresponds to a real-world experimental situation wherein the underlying noise model is completely unknown. This refers to the inverse elasticity problem (Section \ref{sec:elastography}) in this manuscript, where the measurement is obtained from a laboratory experiment.
    \end{enumerate}
    
    \item Prior distribution: we consider five different types of prior distributions that illustrate the diversity in the class of distributions that can be approximated using a WGAN-GP.
\end{enumerate}
For all numerical experiments, we use WGAN-GP \cite{gulrajani2017improved} and train it in TensorFlow \cite{tensorflow2015-whitepaper} by solving the min-max problem in \eqref{eqn:minimax} with the gradient penalty parameter set to $\lambda = 10$. Details about the WGAN architecture and the algorithmic hyper-parameters chosen for each problem can be found in \ref{app:arch}. In all numerical results involving a partial differential equation, the finite element method is used to solve the forward problem in FEniCS \cite{Alnaes2015fenics} and compute the likelihood contribution in  Equation \eqref{eqn:bayes_meas_lsv}. For the inverse radon transform problem (Section \ref{sec:radon_transform}) the discretized forward model is an algebraic equation, and is directly implemented in TensorFlow. We implement the Hamiltonian Monte Carlo (HMC) method with No-U Turn Sampler (NUTS) \cite{Hoffman2014} within Tensorflow Probability \cite{Dillon2017} to sample from the Markov chain (with un-normalized target density given by Equation \eqref{eqn:bayes_meas_lsv}) for inferring the posterior distribution.  In order to compute the gradient of the un-normalized posterior with respect to the latent variable (which is required for an HMC update), we implement a new computational workflow. Within this, the gradient of the forward model is computed using the adjoint method in FEniCS and is passed as an upstream gradient to the TensorFlow computation graph to compute the overall gradient using automatic differentiation. We use an initial step size of 1.0 for HMC and following \cite{Andrieu2008} adapt it based on the target acceptance probability. A burn-in period of 50\% is used for all HMC simulations. These hyper-parameters are selected to ensure that the chains are converged.

A systematic performance comparison (in terms of accuracy and efficiency) of the proposed GAN-based prior with the traditional Bayesian inference method is provided in \ref{app:eff_acc}. In this study, we observe that our proposed algorithm improves efficiency (as measured in effective sample size (ESS))  while retaining similar order of magnitude accuracy as compared to traditional Bayesian inversion method. This highlights the utility of performing Bayesian inference in the lower dimensional latent space. This could be particularly attractive for Bayesian inverse problems arising in science and engineering, which could have $\mathcal{O}(10^4) - \mathcal{O}(10^6)$ inferred parameter dimension (emanating from finer spatio-temporal discretization).

To ensure that the samples produced by the GAN generator satisfy the required physical constraints (such as the positivity of the thermal conductivity in Section \ref{sec:heat_conduction}, density in Section \ref{sec:radon_transform}, and shear modulus in Section \ref{sec:elastography}) we embed these constraints directly into the generator architecture. Specifically, we use \texttt{tanh} activation function at the end of the generator network, which ensures the output of the network is bounded (pixelwise) between -1 and 1. Note that before we train the GAN, the data is transformed (by shifting and scaling) to lie in the range of [-1, 1]. The inverse of this transformation can be used on the output of the generator to get samples with values lying in original (physically consistent) range of the inferred variable, such as thermal conductivity. There are alternate strategies as well to build-in such physical constraints into the neural network, such as by adding physical constraints in the training loss as a (soft) penalty term \cite{Yang2020pigan, Magiera2020ConstraintawareNN, Yang2021EnforcingIC}.

We note that in all numerical results, we compute MAP estimate for our proposed method by evaluating the log of the posterior distribution (up to an additive constant) for every MCMC sample and then designating the sample with the largest value as $\lsv^{\rm MAP}$. Thereafter we treat $\bm{g}^*(\lsv^{\rm MAP})$ as an approximation to $\infv^{\rm MAP}$.

For each problem considered below, we list in Table \ref{tab:dimred} the dimension of the inferred field $\Ninfv$, the dimension of the latent space $\Nlsv$, and the corresponding reduction in dimension measured by the ratio $\Ninfv/\Nlsv$. Note that for some of the problems, we are able to achieve more than two-orders of reduction with the GAN formulation.

\begin{table}[!htbp]
\renewcommand{\arraystretch}{1.5}
\centering
\caption{Dimension reduction using GANs}
\begin{adjustbox}{width=\linewidth}
\begin{tabular}{c c c c c c}
\toprule
Inverse problem & \multicolumn{3}{c}{Inverse Heat Conduction} & Inverse Radon transform & Elastography \\
\cmidrule{2-4}
Dataset &   Rectangular &   MNIST & \begin{tabular}[c]{@{}c@{}}Cahn-Hilliard\\ microstructure\end{tabular} & \begin{tabular}[c]{@{}c@{}}Shepp-Logan\\ phantom\end{tabular} & \begin{tabular}[c]{@{}c@{}}Circular\\ phantom\end{tabular} \\
\begin{tabular}[c]{@{}c@{}}Inferred field\\ dimension ($\Ninfv$) \end{tabular}
  & 784 & 784 & 4096 & 16388 & 3136\\
\begin{tabular}[c]{@{}c@{}}Latent space\\ dimension ($\Nlsv$) \end{tabular} &
 5 & 100 & 100 & 100 & 100 \\
\begin{tabular}[c ]{@{\bfseries}c@{}}Dimension \\ reduction ($\bm{\Ninfv/\Nlsv}$) \end{tabular} &
 $\bm{\approx 157}$ & $\bm{\approx 8}$ & $\bm{\approx 41}$ & $\bm{\approx 164}$ & $\bm{\approx 31}$\\ 
\bottomrule
\end{tabular}\label{tab:dimred}
\end{adjustbox}
\end{table}



\subsection{Inverse heat conduction: coefficient inversion} \label{sec:heat_conduction}
This is a non-linear coefficient inversion problem for an elliptic PDE which arises in many fields such as subsurface flow modeling \cite{Iglesias2013EvaluationOG}, electrical impedance tomography \cite{kaipio2006statistical}, and inverse heat conduction \cite{Kaipio2011}. We focus on the heat conduction problem where the goal is to infer the point-wise thermal conductivity distribution, given partial and noisy measurement of temperature. For this problem the forward model is described by the steady-state heat conduction equation:
\begin{equation}\label{eq:conductivity}
\begin{aligned}
-\nabla \cdot (\kappa (\bm{s}) \nabla u (\bm{s})) &= b(\bm{s}), & \qquad & \bm{s} = (s_1, s_2) \in \Omega & \\
u (\bm{s}) &= 0, & \qquad &\bm{s} = (s_1, s_2) \in \partial \Omega &
\end{aligned}
\end{equation}
where $\Omega \subset \Ro^2$ is a square domain with length = 1 unit, and $b(\bm{s}) = 10^3$ denotes the heat source. The goal is to infer the posterior QoIs of conductivity field $\kappa$, given a noisy, and potentially partial, measurement of temperature field $u$. The nodal values of the temperature field are stored in the vector $\bm{y}$ and those of the conductivity field are stored in the vector $\infv$.  

We compute different QoIs (such as mean, standard deviation etc.) with respect to the posterior distribution using the method proposed in Section \ref{sec:GAN_priors} and expressions for QoIs given in Section \ref{sec:QOI}. Reference QoIs are calculated using random sampling.

We consider three different types of datasets to train the prior for thermal conductivity.
For each dataset, we first train a WGAN-GP to learn the prior distribution using samples from the training set. We then select samples of conductivity from the test set and solve the direct problem to determine the temperature field. We add uncorrelated Gaussian noise to this field and then use all, or a portion, of this field to solve the inverse problem. Finally, we compare the inferred results with the known conductivity field.

We note that finding the optimal latent space dimension of a GAN is still an open research question. For the datasets with the known intrinsic dimension of the underlying manifold (such as the Rectangular dataset considered below), the choice of the latent space dimension of a GAN can be guided by this intrinsic dimension. However, for other datasets (such as MNIST, Cahn-Hilliard microstructure, etc.), where such intrinsic dimension is unknown, it is non-trivial to find the optimal latent space dimension. In this manuscript, we have selected $\Nlsv$=5 for the Rectangular dataset (based on intrinsic dimension) and $\Nlsv$=100 for all other problems (based on empirical evidence and prior numerical study in the literature \cite{Radford2016UnsupervisedRL}). Interested readers can refer to \cite{Marin2021} for a more systematic study on the effect of latent space dimension on GAN performance.

\paragraph{Rectangular dataset}\label{subsec:rect_data}

\begin{wrapfigure}{r}{0.35\textwidth}
  \begin{center}
   \vspace{-2em}
    \includegraphics[width=0.3\textwidth]{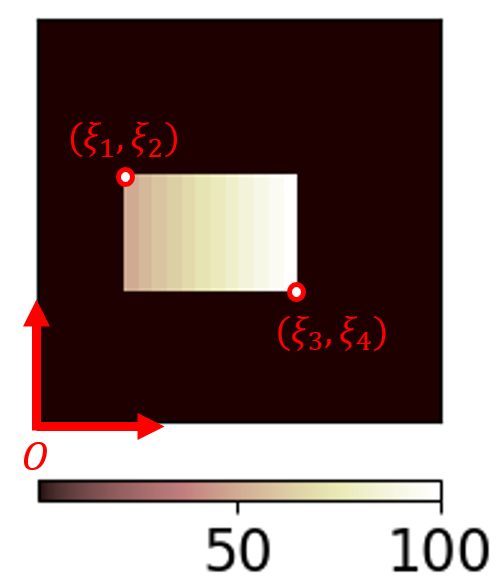}
  \end{center}
  \vspace{-1.7em}
  \caption{Parametric description of the rectangular dataset.}
  \vspace{-1.25em}
  \label{fig:rect_parametric}
\end{wrapfigure}

We first consider a simple dataset so that the predicted QoIs can be compared against their reference values. The conductivity has a uniform background value of unity and a rectangular inclusion whose top-left and bottom-right corners are located at  ($\xi_1, \xi_2$) and ($\xi_3, \xi_4$), respectively, where $\xi_1, \xi_4 \sim \mathcal{U}[0.2, 0.4]$ and $\xi_2$, $\xi_3 \sim \mathcal{U}[0.6, 0.8]$. The thermal conductivity within the inclusion varies linearly from 50 on the left edge to 100 on the right edge. All fields are represented on a $28 \times 28$ Cartesian grid. Figure \ref{fig:rect_parametric} shows parametric description of this dataset.

A dataset of $10^4$ samples of conductivity is generated by considering different realizations of the four $\bm{\xi}$ parameters. Representative samples from this dataset are shown in Figure \ref{fig:priors}(a), and those generated by a WGAN-GP trained using this dataset are shown in Figure \ref{fig:priors}(b). As can be observed from Figure \ref{fig:priors} the generated samples are qualitatively similar to the true samples. 

\begin{figure}[!htbp]
     \centering
        \includegraphics[width=0.8\textwidth]{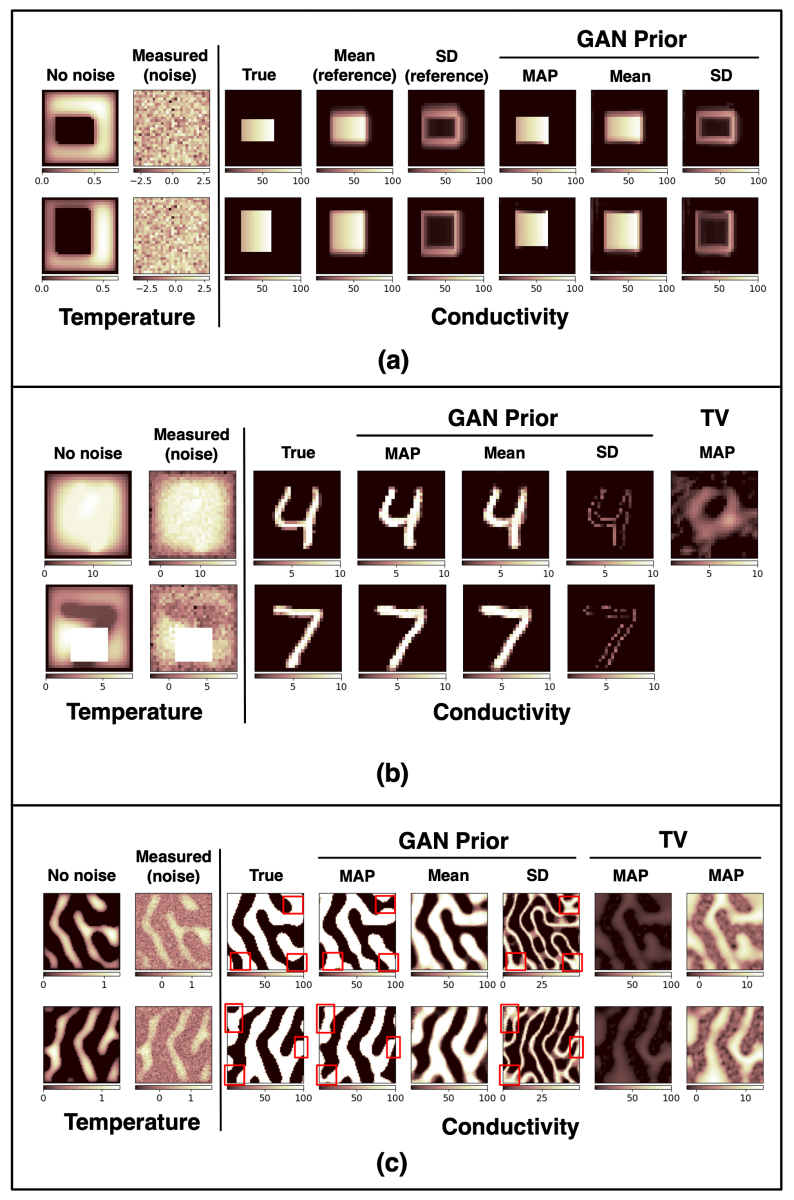}
    \caption{Results for the inverse heat conduction problem. (a) Rectangular dataset. (b) MNIST dataset. (c) Microstructure dataset.}
   \label{fig:cond_combined}
\end{figure}

Images of two test samples of the conductivity field, along with the corresponding temperature field are shown in Figure \ref{fig:cond_combined}(a). The noisy temperature field (with $p_\eta = \mathcal{N}(\bm{\mu}=\bm{0}, \bm{\Sigma}=\sigma_{\rm meas}^2 \mathds{1}$), where $\sigma_{\rm meas}^2= 1$, and $\sigma_{\rm meas}^2/||\bm{f}(\bm{\infv})||_2^2$ = 8.8 and 15.5, respectively for two test examples shown in Figure \ref{fig:cond_combined}(a)) is used as the measurement. The algorithm described in  Figure \ref{fig:schematic} is used to infer the conductivity field, and compute the QoIs. These include the maximum a-posteriori (MAP) estimate, the mean, and the point-wise standard deviation (Figure \ref{fig:cond_combined}(a)). This figure also shows the true conductivity, and the ``reference'' mean and standard deviation fields, which are computed by sampling the space of the four parameters, $\bm{\xi}$. 
This comparison with the reference QoIs is possible only because the explicit four-dimensional representation for this conductivity field is known. The inferred MAP corresponds to the best guess for the true conductivity field. By comparing it with the true field, we conclude it is quite accurate even in the presence of significant measurement noise. In the mean field we observe that the edges of the inclusion are blurred thereby indicating the uncertainty in determining its boundaries. This is further highlighted in the plot of the standard deviation (SD) field which is elevated along the edges. A comparison of the inferred QoIs with their true counterparts verifies the accuracy of the proposed method.


\paragraph{MNIST dataset}\label{subsec:mnist}
The conductivity field is constructed from the MNIST dataset \cite{lecun2010mnist} of handwritten digits by linearly scaling the image intensity to be between 1 and 10 units. All fields are represented on a $28 \times 28$ Cartesian grid.
%
Representative samples from the true prior density of the conductivity field are shown in Figure \ref{fig:priors}(e). $5 \times 10^4$ images from this dataset are used to train the WGAN-GP, and samples from the trained generator are shown in Figure \ref{fig:priors}(f). Again, we note that the generated samples are qualitatively close to the true samples.

Measured temperature fields are generated by solving the direct problem for a conductivity field selected from the test set and then adding Gaussian noise to it; that is, with $p_\eta = \mathcal{N}(\bm{\mu}=\bm{0}, \bm{\Sigma}=\sigma_{\rm meas}^2 \mathds{1}$), where $\sigma_{\rm meas}^2= 1$, and $\sigma_{\rm meas}^2/||\bm{f}(\bm{\infv})||_2^2$ = 0.008 and 0.12, respectively for two test examples shown in Figure \ref{fig:cond_combined}(b). 
Results are reported in Figure \ref{fig:cond_combined}(b).
The top row in this panel shows results for a measurement where the temperature field is known in the entire domain, and the bottom row shows results for a partial measurement. In both cases, by comparing the true conductivity field with the inferred MAP, we discern the ability of the proposed method in performing accurate inference with complex prior data. This is especially evident when comparing the MAP with the inferred result from a total variation diminishing (TV) prior, which is one of the more popular regularization methods for solving inverse problems (see \ref{app:TV} for details of TV algorithm). We note that we use TV regularization here as a benchmark to show the comparison of our data-driven method against the more generic and analytical regularizer method used traditionally in inverse problems. The use of more informative prior/regularizer could lead to better reconstruction results. The mean and the standard deviation fields illustrate that the inference is most uncertain along the interface between the regions of high and low values of conductivity. 




An inference algorithm in Bayesian setting should perform consistently across different noise levels in measured field. This ensures that the obtained results are robust and reliable. To investigate this we study the effect of noise level on reconstruction results obtained by our algorithm and compare it against reconstruction obtained from TV regularization. 
\begin{figure}[!htbp]
    \centering
    \includegraphics[width=0.9\textwidth]{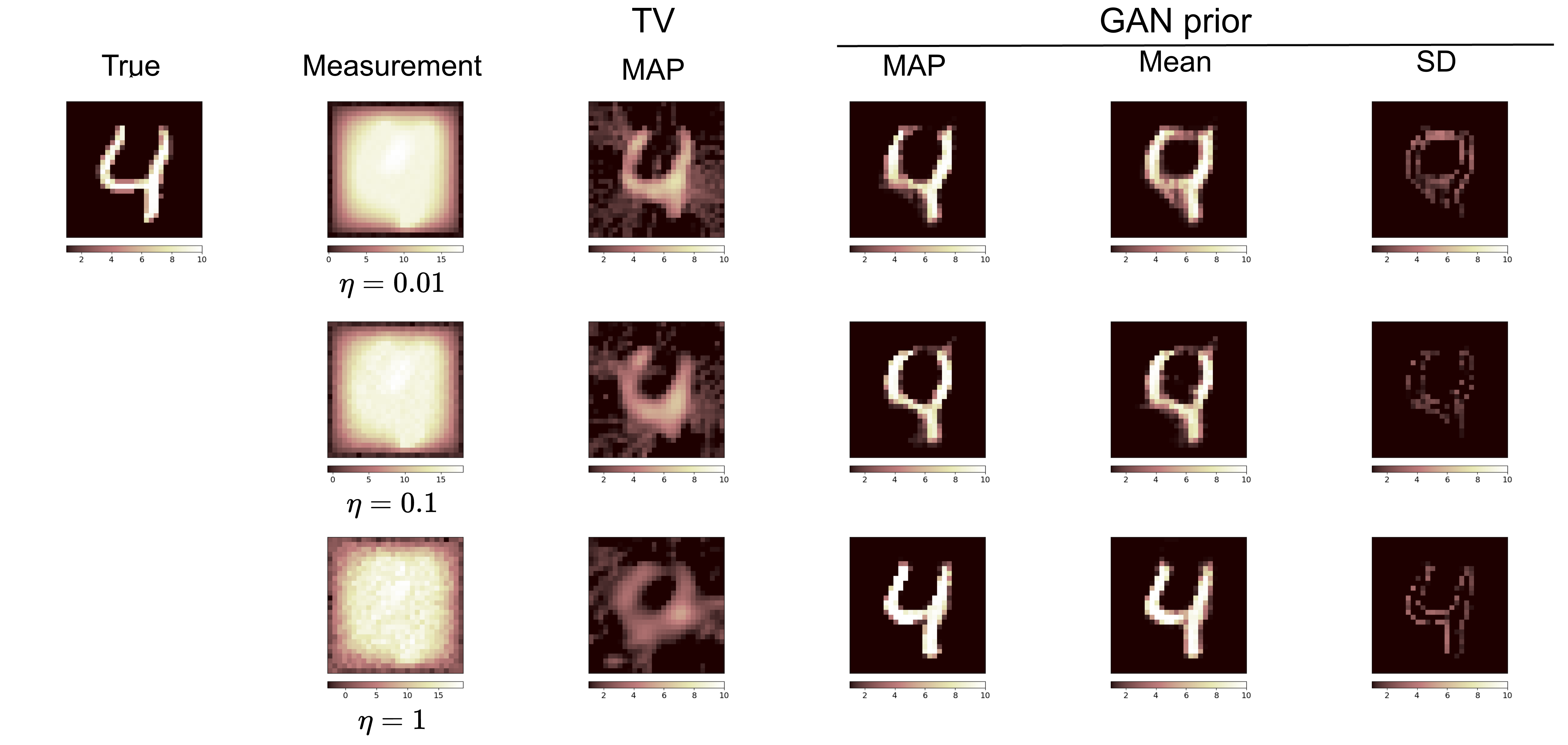}
    \caption{Effect of measurement noise on reconstruction results. \textit{First column}: true conductivity field from test set. \textit{Second column}: temperature measurements with different level of additive Gaussian noise.  \textit{Third column}: reconstruction results obtained by solving deterministic optimization problem with TV regularization. \textit{Fourth, Fifth, and Sixth column}: MAP, Mean, and Standard deviation estimate obtained from our proposed GAN prior method. }
    \label{fig:tv_gan_compare}
\end{figure}

Figure \ref{fig:tv_gan_compare} shows the results of this study showcasing the effect of measurement noise on reconstruction quality. The second column in Figure \ref{fig:tv_gan_compare} shows then temperature field corrupted with measurement noise. For this study we consider additive Gaussian noise as noise model with three different level of noise variance: $\sigma_{\rm meas}^2= 0.01, 0.1, \text{and } 1.0$. The third and the fourth column shows the MAP estimate obtained using TV regularization and our algorithm respectively. The final two columns show the mean and the standard deviation estimates respectively obtained using our proposed algorithm. As can be observed, the GAN-based prior performs significantly better than TV at all noise levels. Furthermore, the performance of this method stays consistent even at very high noise level ($\eta$ = 1), whereas the performance of the TV method degrades significantly. This should not be surprising, since our method learns prior directly from data and hence uses prior density which is more close to ``true'' prior density, whereas TV method uses pre-defined analytical form of regularization promoting a piece-wise constant solution, which might not necessarily be the optimal choice and hence not the strong prior. This highlights a key feature of our proposed method: using a prior density, which is learned directly from data, leads to stronger priors which enable better posterior approximation even in the presence of high measurement noise. We note that the important hyperparameters of the TV method, such as the regularization parameter, were optimally tuned using L-curve criteria \cite{Calvetti2000} to ensure fair comparison. In \ref{app:TV} we provide these L-curves for all three noise levels.



\paragraph{Two-phase material microstructure} 
We now consider a complex dataset of  material microstructure for a bi-phase material.
Multiple realizations of the the conductivity field for this dataset are obtained by solving the Cahn-Hilliard equation with randomly selected initial conditions. 
All fields are represented on a $64 \times 64$ Cartesian grid. Figure \ref{fig:priors}(c) shows four representative samples from the training set, and Figure \ref{fig:priors}(d) shows four samples from the generator of the trained WGAN-GP.

The results of the inference problem are shown in Figure \eqref{fig:cond_combined}(c). These include the noisy temperature field ($\sigma_{\rm meas}^2= 0.05$, and $\sigma_{\rm meas}^2/||\bm{f}(\bm{\infv})||_2^2$ = 0.35 and 0.39, respectively for two test examples shown in Figure \ref{fig:cond_combined}(c)), which serves as the measurement, the MAP, the mean and the standard deviation of the reconstructed field. When comparing the MAP with the true conductivity field, we observe that the former is able to capture most of the streaky structure in the true field. However, there are local regions (indicated by red rectangles) where the MAP and the true field differ. These occur near the boundaries, where temperature is prescribed through the boundary condition, and therefore is only weakly influenced by the conductivity. Remarkably, in these regions the value of the standard deviation is elevated, clearly indicating that the uncertainty in the prediction is high, and therefore the confidence in the MAP field ought to be low. This ability to make quantitative assessments of the confidence in predictions is one of the main motivations for using probabilistic methods for solving inverse problems. A comparison of the MAP obtained using the GAN prior with the TV prior shows that the latter severely under-predicts the contrast in the conductivity field and produces oscillations that are absent from the true field. We have shown two images of the TV MAP. The image on the left is plotted using the scale that was used for the true, and the GAN prior (MAP and Mean) images, while the image on the right is plotted using a narrower scale and highlights the oscillations in the TV MAP.

\subsection{Inverse heat conduction : source identification}
\label{sec:source_inversion}

In this section we consider a different inverse problem that arises in steady state heat conduction - one of inferring the source term from a measurement of the temperature. This inverse problem plays a critical role in the design and development of thermal equipment and it has garnered significant research attention \cite{Cannon1986, Cannon1998structural, Su2001, Cheng2010}.

The problem of interest is given by,
\begin{equation}\label{eq:conductivity_source}
\begin{aligned}
-\nabla \cdot (\kappa (\bm{s}) \nabla u (\bm{s})) &= b(\bm{s}), & \qquad & \bm{s} = (s_1, s_2) \in \Omega & \\
u (\bm{s}) &= 0, & \qquad &\bm{s} = (0, s_2) \cup (L, s_2) \in \partial \Omega_g & \\
\frac{\partial u(\bm{s})}{\partial \bm{s}}  &= 0, & \qquad &\bm{s} = (s_1, 0) \cup (s_1, L) \in \partial \Omega_h & 
\end{aligned}
\end{equation}
where $\Omega \subset \Ro^2$ is a square domain with length = 1 unit (discretized on a Cartesian grid of size $28 \times 28$), and $\kappa(\bm{s})$ denotes thermal conductivity, which is assumed to be a constant and equal to 0.01 units. The goal here is given a noisy measurement of temperature field $u$, infer the posterior QoIs of source field $b$. The nodal values of the temperature field are stored in the vector $\bm{y}$ and those of the source field are stored in the vector $\infv$.  

\begin{figure}[!htbp]
    \centering
    \includegraphics[width=0.9\textwidth]{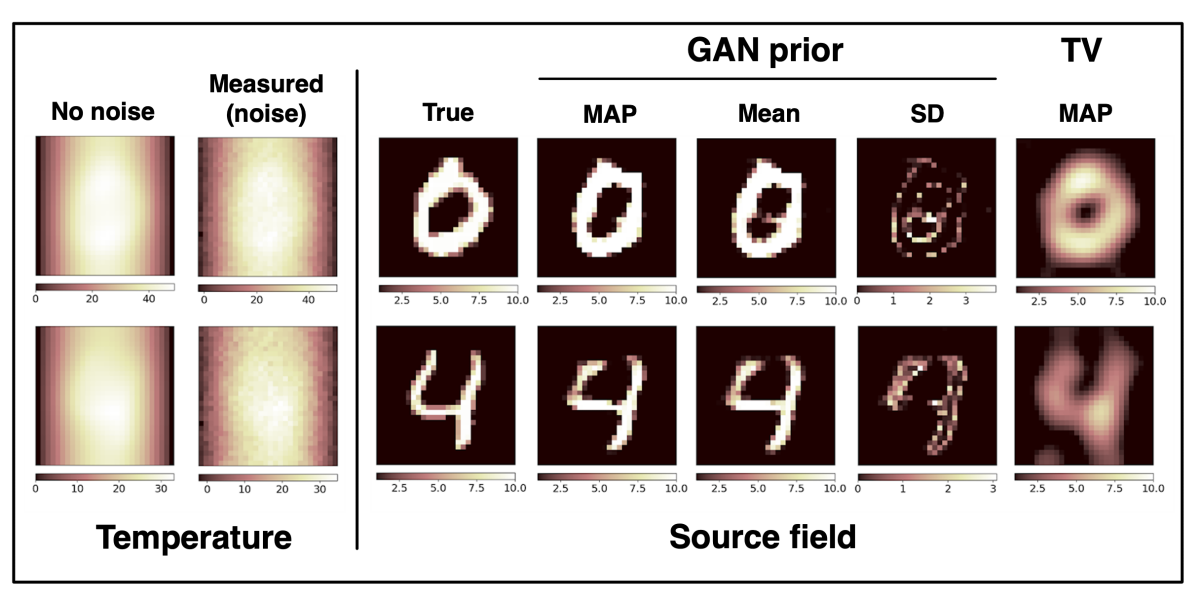}
    \caption{Source identification from noisy temperature measurement (MNIST dataset). }
    \label{fig:source_mnist}
\end{figure}

In this numerical experiment we consider the MNIST dataset as the prior distribution for the source field. 
Figure \ref{fig:source_mnist} shows result for this problem for two test samples. Here the left panel shows the temperature field - with and without additive Gaussian noise ($\sigma_{\rm meas}^2= 1$). The results obtained for the noisy temperature measurement field are shown in the right panel. Specifically, we show the MAP, mean, and standard deviation estimates and compare them against a result obtained using TV regularization. As can be observed, the TV reconstruction is able to capture the overall structure of the solution but fails to capture the sharp transitions in the source field, whereas the method based on the GAN-prior is significantly better and closer to the true field. The standard deviation is elevated at the interface between the high and low intensity regions, which is consistent with what might be expected in this case. 

\subsection{Inverse Radon transform}
\label{sec:radon_transform}

Next, we demonstrate the efficacy of the proposed Bayesian inference strategy in the context of inverse Radon transforms, which are routinely used in computerized tomography (CT) \cite{natterer2001}. We restrict our discussion to two-dimensional inputs (slices) scanned using one-dimensional line-Radon transforms. Given a function $\rho: \Omega \subset \Ro^2 \mapsto \Ro$, the Radon transform is given by
\begin{equation}\label{eqn:1dradon}
\mathcal{R}(\rho;t,\psi) := \int_{\ell_{t,\psi}} \rho d \ell,
\end{equation}
where $\ell_{t,\psi}$ is the line through $\Omega$ inclined at an angle $\psi$ and at a signed-distance of $t$ from the center of $\Omega$. For the current problem, we consider the Shepp-Logan head phantom \cite{toft_1996}, which 
%
is composed of a union of ten ellipses with each ellipse having a different but constant density. The $k$-th ellipse $E_k$ is centered at $(r_k,s_k)$, with semi-axis lengths $a_k$, $b_k$, angle of inclination $\alpha_k$ and density $\rho_k$. The precise values of these parameters having been taken from \cite{toft_1996} and are also tabulated in \ref{app:phantom}. To generate training and test samples, the original parameters are randomly sampled to obtain a perturbed ellipse $\tilde{E}_k$ as follows:
\begin{equation}\label{eqn:sl_params}
\begin{aligned}
\tilde{r}_k &= r_k + 0.005\omega_k^1, \quad \tilde{s}_k = s_k + 0.005\omega_k^2, \quad \tilde{a}_k = a_k + 0.005\omega_k^3, \\
\tilde{b}_k &= b_k + 0.005\omega_k^4, \quad \tilde{\alpha} = \alpha_k + 2.5\omega_k^5, \quad \tilde{\rho}_k = \rho_k + 0.0005\omega_k^6.
\end{aligned}
\end{equation}
The random parameters in \eqref{eqn:sl_params} are sampled uniformly as $\omega_k^i \sim \mathcal{U}[-1,1]$. The density $\rho$ for the perturbed Shepp-Logan phantom is given by
\begin{equation}\label{eqn:pert_sl}
\rho(r,s) = \max \left(0,\min\left(1,\sum_{k=1}^{10} \tilde{C}_k(r,s)\right)\right), \quad \tilde{C}_k(r,s) = \begin{cases} \rho_k  & \quad \text{if } (r,s) \in \tilde{E}_k\\
0 & \quad \text{otherwise} \end{cases} 
\end{equation}
The clipping in \eqref{eqn:pert_sl} ensures that the total density at any point in the phantom ranges from 0 (air cavity) to 1 (bone). 

The discrete phantom $\infv$ obtained by evaluating \eqref{eqn:pert_sl} on a grid of $128 \times 128$. This phantom image is further transformed as $\infv \leftarrow \mathcal{F}(\infv; n,m,\beta)$ to generate a single training (or test) sample. The transformation $\mathcal{F}$ translates $\infv$ by $n$ pixels in the horizontal direction, $m$ pixels in the vertical direction and rotates by an angle $\beta$, where 
\begin{equation}\label{eqn:sl_trans}
n,m \sim \mathcal{U}\{-8,-7,...,7,8\}, \quad \beta \sim \mathcal{U}[-20^\circ,20^\circ].
\end{equation}
Four samples of the discrete phantom are shown in Figure \ref{fig:priors}(g). Note that the randomness in generating the samples of $\infv$ is attributed to the perturbation in the phantom parameters, as well as the random affine transformation of the discrete phantom.

\begin{figure}[!htpb]
\begin{center}
\includegraphics[width=\textwidth]{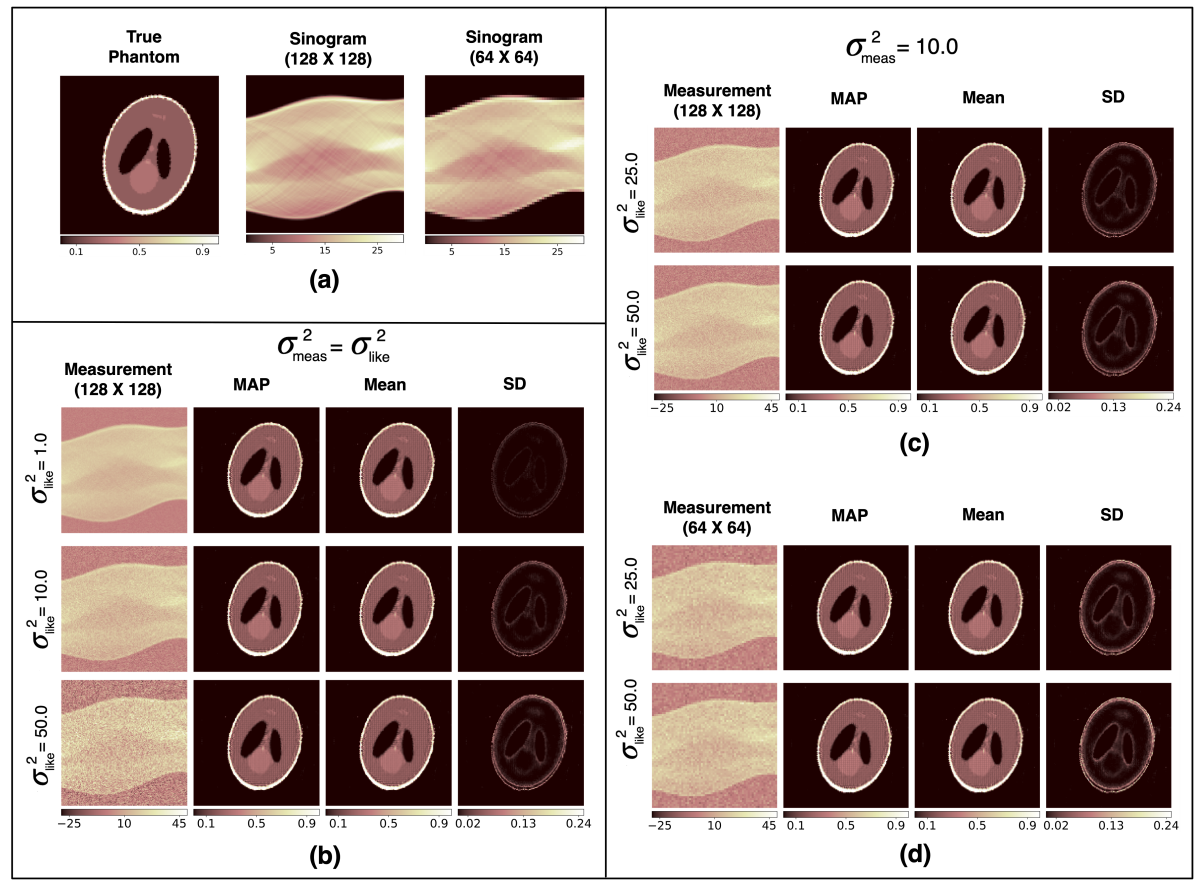}
\caption{Results for the inverse Radon transform problem. (a) True (test) sample of the Shepp-Logan phantom and the corresponding sinograms at two different resolutions. (b) Inference results with measurements at the finer scale and when the noise level of the measurement is known precisely. (c) Inference results with measurements at the finer scale and when the noise level of the measurement is unknown. (d) Inference results with measurements at the coarser scale and when the noise level of the measurement is unknown.}
\label{fig:radon}
\end{center}
\end{figure}

The direct map for the problem is given by
\begin{equation}\label{eqn:ct_fmap}
\begin{aligned}
    \fmap(\infv) &= \meas \in \Ro^{K \times K},  \\ 
    y_{i,j} &= \mathcal{R}^h(\infv;t_i,\psi_j), \\
    t_i &= \frac{i}{K}, \quad \forall \ 1\leq i \leq K,\\
    \psi_j &= \frac{j}{\pi}, \quad \forall \ 1\leq i \leq K,
\end{aligned}
\end{equation}
where the discrete Radon transform $\mathcal{R}^h$ along each of the $K^2$ lines is computed using a quadrature.    
The resulting $K \times K$ image is known as a sinogram, an example of which is shown in Figure \ref{fig:radon}(a). 

A WGAN-GP is trained on a dataset with $10^4$ discrete phantom images. Four realizations from the learned prior are shown in Figure \ref{fig:priors}(h); they are qualitatively similar to the true samples (shown in Figure \ref{fig:priors}(g)). We choose a sample from the test set (see Figure \ref{fig:radon} (a)), and add Gaussian noise $\mathcal{N}(\bm{0},\sigma^2_{\text{meas}} \mathbb{I})$ to the corresponding sinogram to generate the noisy measurement $\nmeas$. We then infer the posterior QoIs using the algorithm in Figure \ref{fig:schematic}, with the likelihood term in \eqref{eqn:bayes} modelled using $\mathcal{N}(\bm{0},\sigma^2_{\text{like}} \mathbb{I})$. Three different scenarios are considered for the inference process:

\paragraph{Case 1: $\sigma^2_{\text{meas}} = \sigma^2_{\text{like}}$} We assume complete knowledge of the noise model while constructing the likelihood term. Setting $K=128$, which we term as the fine resolution, the inference results for varying levels of noise are shown in Figure \ref{fig:radon}(b). For each value of $\sigma_{\text{meas}}$, we recover the MAP estimate, mean and pixel-wise standard deviations. Even for the highest noise level in the measurement, we get an excellent recovery of the phantom, with a relatively low variance concentrated along the boundaries of the ellipses.

\paragraph{Case 2: $\sigma^2_{\text{meas}} \neq \sigma^2_{\text{like}}$} We work with the same level of measurement resolution as in Case 1 and set $\sigma^2_{\text{meas}} = 10.0$. However, we assume that we do not know the the variance of the noise.
The inference results shown in Figure \ref{fig:radon}(c) with different values of $\sigma^2_{\text{like}}$ indicate a good recovery of the phantom. This case corresponds to the scenario where we only know the form of the distribution of the noise model, but do not know the precise values of its parameters.

\paragraph{Case 3: $\sigma^2_{\text{meas}} \neq \sigma^2_{\text{like}}$ with partial observation} We consider the same situation as Case 2, but with the measurements taken at a coarser level, i.e., $K=64$. The inference results are shown in Figure \ref{fig:radon}(d). Note that the variance along the phantom boundaries is slightly elevated compared to case 2, which is expected since the uncertainty in the results should increase with a decrease in the resolution of observed data.

\subsection{Elasticity imaging} \label{sec:elastography}
In this section we consider the inverse problem with experimentally measured data with application to elasticity imaging. Elasticity imaging is a technique for inferring mechanical properties of tissue from displacement data \cite{barbone2010review,doyley2014elastography}. The direct problem is given by the equations of the balance of linear momentum for an incompressible linear elastic solid in a state of plane stress, 
 \begin{equation}\label{eq:elasto}
     \begin{aligned}
         \nabla \cdot \bm{\sigma} &= \bm{0} &&\text{in $\Omega$} \\
         \bm{u} &= \bm{u_D}  &&\text{on $\Gamma_D$} \\
         \bm{\sigma} \cdot \bm{n} &= \bm{\tau}  &&\text{on $\Gamma_N$} 
     \end{aligned}
 \end{equation}
where $\bm{\sigma} = 2 \mu (\nabla^{s}\bm{u} + (\nabla \cdot \bm{u})\mathds{1})$ is the stress tensor, $\mu$ is the shear modulus, and $\bm{u}$ is the displacement field. In the problem considered in the study, the domain $\Omega$ is a rectangle of dimension $34.608 \times 26.287\ mm$. On the vertical edges of the rectangle, a zero traction boundary condition is prescribed, whereas on the horizontal edges the vertical displacement component is set equal to its measured vertical displacement value and horizontal traction is set to zero. The nodal values of the vertical displacement and the nodal values of the shear modulus constitute the measurement vector $\bm{y}$, and inferred vector $\infv$, respectively. Here the inverse problem is one of determining the shear modulus given the vertical displacement field. 

The data used for solving the elasticity imaging inverse problem was obtained from a physical experiment on a tissue-mimicking phantom manufactured from a mixture of gelatin, agar, and oil \cite{Pavan2012}. The physical specimen comprised of a stiff spherical inclusion embedded in a softer background. This phantom was gently compressed and the vertical component of the displacement field in the central plane of phantom was measured using ultrasound. This measured field was restricted to a rectangular physical domain (of size $34.608 \times 26.287\ mm$) and represented on a $56 \times 56$ finite element grid. This same grid was also used to represent the shear modulus. 

In solving the inverse problem prior information about the shear modulus field was assumed. In particular, it was assumed that the field comprised a uniform circular inclusion embedded within a uniform background. The center of the inclusion and the contrast in the shear modulus value between the inclusion and the background, were treated as random parameters. A set of $6 \times 10^3$ realizations of the shear modulus field was obtained by uniformly sampling these parameters, and this set was then used to train a WGAN-GP to represent the prior distribution. 
Additional details about this dataset and the values of randomly sampled parameters are provided in  \ref{app:elsto_phantom}.

\begin{figure}[!htbp]
     \centering
         \includegraphics[width=\textwidth]{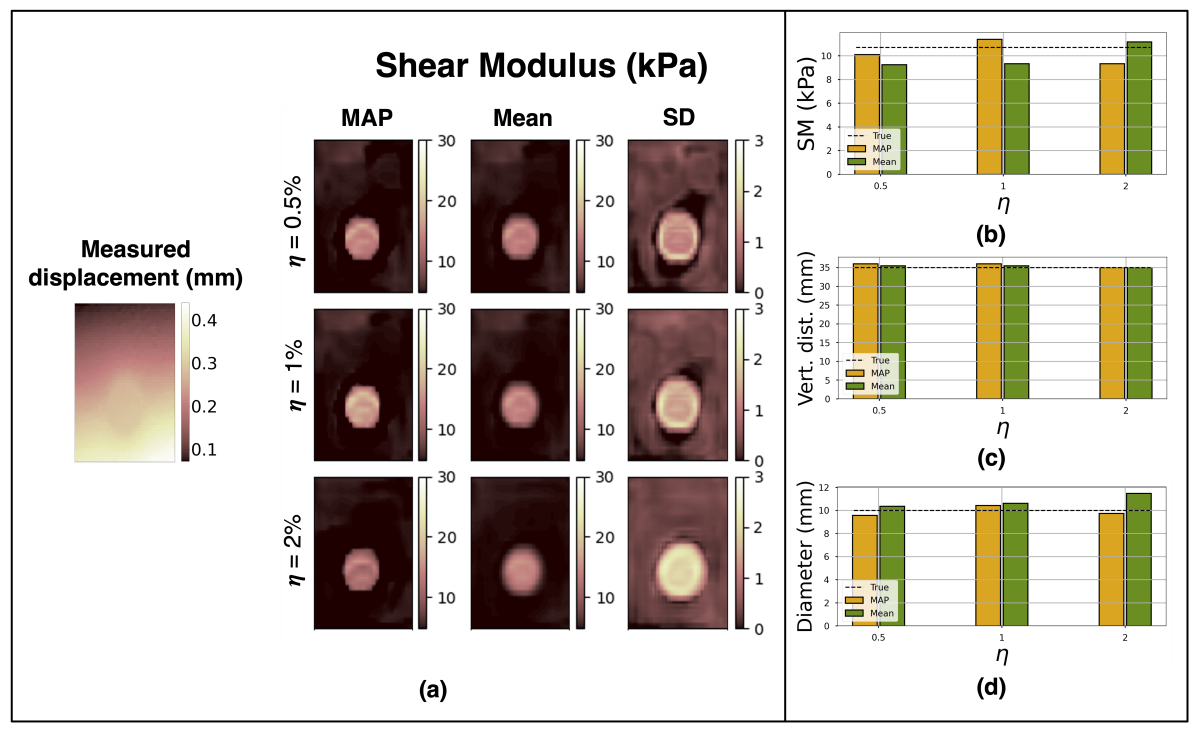}
    \caption{Results for the Elasticity Imaging problem. (a) From left to right: vertical displacement field measured in the phantom, MAP, Mean and Standard Deviation field obtained by solving the inverse problem. Each row corresponds to a different value of assumed variance in noise. (b) Shear modulus of the inclusion evaluated using the MAP and Mean fields. (c) Vertical coordinate of the center of the inclusion evaluated using the MAP and Mean fields. (d) Diameter of the inclusion evaluated using the MAP and Mean fields.}
    \label{fig:elasto}
\end{figure}

The measured vertical displacement field is shown in Figure \ref{fig:elasto}(a). Since this field was obtained experimentally, the true model for its noise, which is needed to evaluate the likelihood term in the posterior distribution, was not known. So, the noise (likelihood) distribution has to be \textit{assumed}. It was assumed to be isotropic Gaussian, with three different values of variance used for evaluating the posterior distribution (see Figure \ref{fig:elasto}). In each case, the MAP field revealed a circular inclusion with an elevated modulus, a mean field where the edge of the inclusion was blurred, and a standard deviation field that was elevated within the inclusion. The higher value of the standard deviation within the inclusion is consistent with the fact that in the stiffer regions of the phantom, a perturbation in the shear modulus leads to smaller changes in the displacement, and therefore for a spatially uniform measurement error, greater uncertainty in predictions. A quantitative comparison of the reconstructed parameters and their true counterparts is also shown in Figure \ref{fig:elasto}. Here the MAP and the mean fields obtained from the reconstructions corresponding to the three different assumed noise levels were used to compute the shear modulus of the inclusion (b), the vertical distance of its center from the top (c), and its diameter (d), which were compared with experimental measurements. All values were found to be within 10\% of their experimental counterparts. 




\section{Conclusions}\label{sec:conclusions}
Bayesian inference is a popular framework for solving inverse problems. However, it can be challenging to use when inferring vectors of large dimensions, and in cases where prior information is available through previously acquired samples or historical data. In this manuscript we use GANs as priors to address both these challenges. The efficacy of our approach is demonstrated on a wide range of inverse problems. These include problems where (a) the direct problem is governed by differential or integral equations, (b) the prior knowledge for the quantity to be inferred is obtained from different sources including parametric descriptions, natural images, and solutions of equations that govern the evolution of micro-structure, and (c) the likelihood distribution is known to varying degrees. In each case, in addition to evaluating the most likely value for the inferred field, we compute QoIs like the mean and standard deviation fields. These fields provide guidance to the end user by identifying regions of high uncertainty. For example, in the elasticity imaging problem considered in this paper, we observe that the largest uncertainty occurs in regions with elevated shear modulus. This is precisely where the measured displacement is most insensitive to the modulus. Similarly, in the heat conduction problem with micro-structure, the largest uncertainty occurs along the boundaries of the domain where the temperature values are determined by primarily the boundary conditions and not the thermal conductivity field. This ability to efficiently quantifying uncertainty in high-dimensional inferred fields is critical in many applications, where high-stake decisions are made based on the output of the inference algorithm or where this uncertainty information is leveraged in downstream tasks such as forward uncertainty propagation, sensitivity analysis, rare event prediction, and optimal experimental design. 

We end with a discussion of the limitations of the approach discussed in this manuscript. First, it is applicable when a sufficiently large data set of previously acquired samples or historical data of the inferred field is available. While this may be available for some cases (like a dataset of material microstructure, subsurface flow and groundwater channels, bathymetry in hydrology, and medical images) and could be generated for others (via additional experiments, field studies, simulations or data augmentation), there are scenarios where this may not be possible. For these cases, other means of describing the prior ought to be considered. Second, using this approach involves training a WGAN-GP in the first phase and an MCMC sampler in the second phase. Both these tasks require careful consideration of the type of problem being solved, and the values of the hyper-parameters to be selected.

\section{Acknowledgements}

The authors acknowledge support from ARO grant W911NF2010050 and from the Airbus Institute for Engineering Research at USC. The authors acknowledge the Center for Advanced Research Computing (CARC) at the University of Southern California for providing computing resources that have contributed to the research results reported within this publication.

\newpage 

\appendix

\section{Efficiency and Accuracy comparison of the proposed GAN-based prior method with traditional Bayesian inference}\label{app:eff_acc}

One of the primary objectives of any good Bayesian inference algorithm is to compute posterior statistics reasonably accurately with a reasonable computation cost. In this section, we have done a systematic study comparing the performance (in terms of efficiency and accuracy) of our proposed GAN-based method with that of the traditional Bayesian inference method. Specifically for a selected problem, we have first performed Bayesian inference in the visible space ($\infv$-space) using a known prior and MCMC method (this is a traditional way of performing Bayesian inference) and computed its accuracy and efficiency (as described below). Then for our method, we first generated samples from the \textit{same} prior to train a GAN, and used this trained GAN-based prior to perform inference in the lower dimensional latent space ($\lsv$-space) using the \textit{same} MCMC method and compute the \textit{same} performance metrics. We observe that our proposed method can achieve significant improvement in efficiency while retaining more or less similar accuracy. Next, we provide the details of the problem.
   
   In order to do a fair and rigorous comparison between traditional Bayesian inference (in the visible space) and our proposed low-dimensional inference method (in the latent space), we have selected a problem which satisfies the following desirable properties:
   \begin{itemize}
       \item The choice of prior and likelihood distribution should be such that the resulting posterior statistics can be analytically computed (for using it as reference statistics for accuracy calculations).
       
       \item Prior distribution should be such that samples can be easily generated from it to train a GAN.
       
       \item There should exist some suitable low-dimensional approximation of the prior distribution to make the notion of dimension reduction meaningful.
       
       \item The target posterior distribution should be non-trivial and challenging, for instance, a highly ill-conditioned posterior density.
   \end{itemize}
   
   Conjugate prior with reasonably complex prior density satisfies all of the above properties and hence in this study, we have considered that as our test problem. Specifically, we consider a strongly ill-conditioned Gaussian (ICG) as our prior distribution with an identity forward map and Gaussian likelihood. This results in an ICG as a posterior distribution for which the posterior statistics can be analytically computed. Furthermore, the posterior has a highly non-trivial geometry making it a benchmark problem of choice to test different MCMC algorithms \cite{Lvy2018GeneralizingHM, hoffman2019neutralizing}. We consider the following scenario: $\infv$-space dimension = 200, $\lsv$-space dimension = 10, i.e. dimensionality reduction by a factor of 20. 
   
   For all test cases, we use Hamiltonian Monte Carlo algorithm for inference. We tune the step size and the number of leapfrog steps of HMC for all test cases such that the acceptance probability is in the target range of 0.6-0.9. We report accuracy in terms of error in the posterior mean and covariance, and efficiency in terms of effective sample size (ESS) per sample which is computed as below:
   
   Let $\{\bm{z_{\tau}}\}_{\tau \leq T}$ be a sequence of correlated samples from an MCMC chain converging to some distribution $p$ with mean $\bm{\mu}$ and covariance $\bm{\Sigma}$. We then define auto-correlation at time $t$ as:
   \begin{equation*}
       s(t) = \frac{1}{trace(\Sigma)(T-t)}\sum_{\tau \leq T-(t+1)} (\bm{z_{\tau}}-\bm{\mu})^T(\bm{z_{\tau+t}}-\bm{\mu}).
   \end{equation*}
   The ESS is then defined as 
   \begin{equation*}
       ESS(\{\bm{z_{\tau}}\}_{\tau \leq T}) := \frac{1}{1+2\Sigma_{t} s(t)}.
   \end{equation*}
   Similar to \cite{nuts_hmc} and \cite{Lvy2018GeneralizingHM}, we truncate the sum when the auto-correlation goes below 0.05.

We choose prior distribution to be multivariate Gaussian with mean vector $\bm{\mu_{prior}} \in \Ro^{200} = [10, 10, \cdots, 10]$ and covariance matrix to be a diagonal matrix with diagonal entry being $[10^3, 10^3, 10^{-3}, 10^{-3}, \cdots, 10^{-3}]$. We use identity forward map and Gaussian likelihood with zero mean and a diagonal covariance matrix with diagonal entry being $[1, 1, 10^3, 10^3, \cdots, 10^3]$. This results in an ill-conditioned Gaussian as a posterior distribution, which is very difficult to sample from as reported in previous studies with similar problems \cite{Lvy2018GeneralizingHM, hoffman2019neutralizing}. We first perform inference in the visible space ($\infv$-space) using HMC algorithm. We ran five different chains with five different random seeds and report the average results with one standard deviation in the following table. For the GAN-prior case, we train a WGAN-GP with the latent dimension=10 using the samples from the prior and perform inference in the latent space ($\lsv$-space). Again we report average results with one standard deviation for five different Markov chains with different random seeds.

\begin{table}[!htbp]
\renewcommand{\arraystretch}{1.5}
\centering
\caption{Comparison of traditional Bayesian inference with GAN-based priors with latent space inference}
\begin{adjustbox}{width=0.65\linewidth}
\begin{tabular}{c c c c}
\toprule
Method & Rel. error in mean & Rel. error in std. dev. & ESS/sample  \\
\midrule
$\infv$-space inference  & \multirow{1}{*}{1.15 $\pm$ 0.45 \%} & \multirow{1}{*}{0.434 $\pm$ 0.15 \%} & \multirow{1}{*}{5.14e-3 $\pm$ 8.86e-3}
 \\
$\lsv$-space inference & 0.66 $\pm$ 0.14 \% & 3.5 $\pm$ 0.62 \% & 7.5e-1 $\pm$ 2.3e-1 \\

\bottomrule
\end{tabular}\label{tab:case2}
\end{adjustbox}
\end{table}

  As can be observed from Table \ref{tab:case2}, the relative error in posterior statistics is more or less of the same order of magnitude in both methods, but the GAN-prior method with latent space inference mechanism remarkably outperforms traditional Bayesian inference method in terms of efficiency. In high dimensions with extremely narrow posterior distribution, typically the HMC algorithm faces challenges in converging \cite{betancourt2015hamiltonian}. As a result one has to take extremely small Hamiltonian steps (i.e. product of step size and number of leapfrog steps) to ensure convergence. This results in highly correlated samples and low ESS, whereas with the GAN-prior method the posterior distribution is low-dimensional (a factor of 20 reduction) and relatively well behaved, which helps in achieving higher efficiency.

\section{WGAN-GP architecture, MCMC sampling, and training hyperparameters \label{app:arch}}
\begin{figure}[!htbp]
\begin{center}
\subfigure[Architecture A (rectangular)]{\includegraphics[width=0.45 \linewidth]{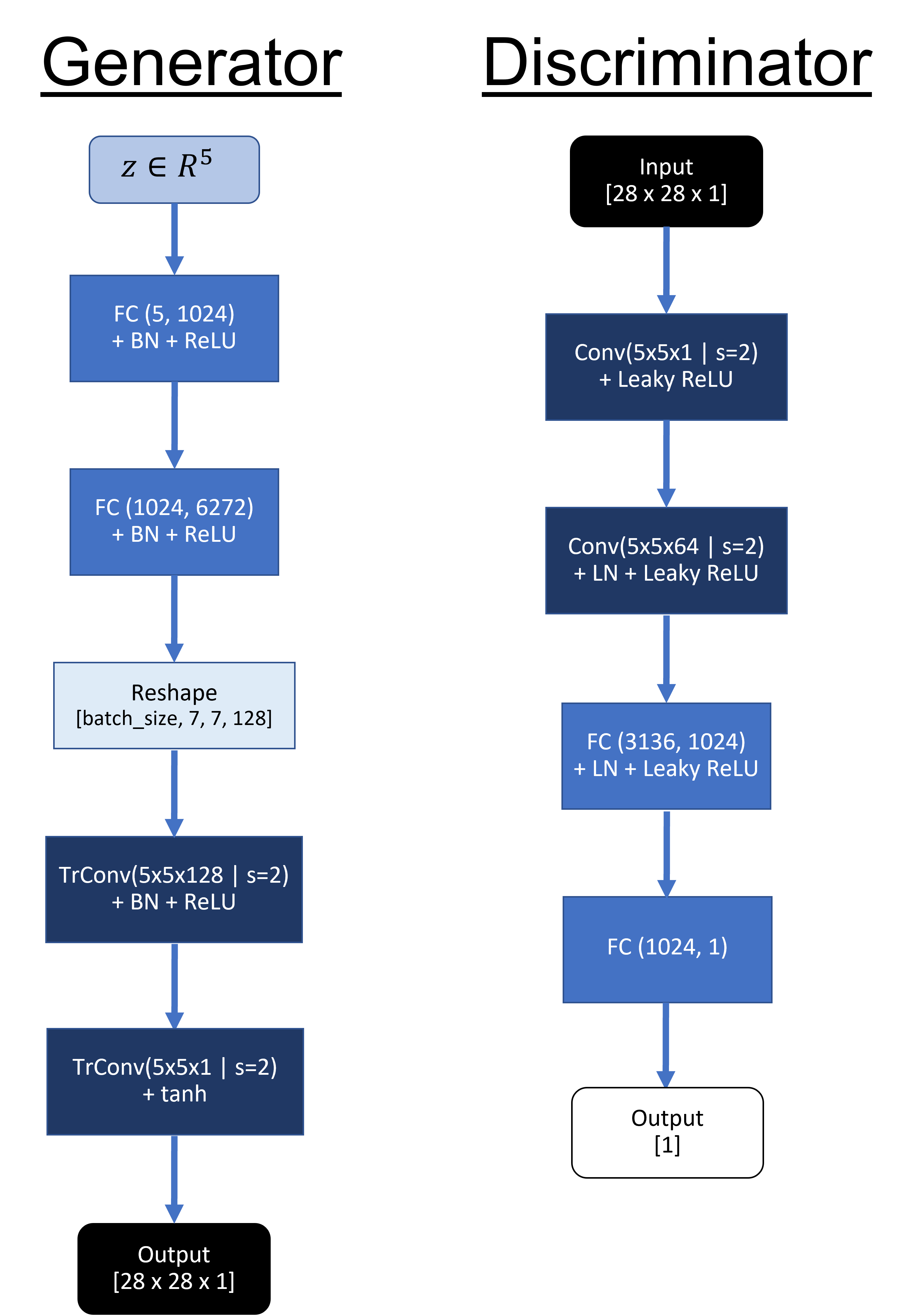}} 
\subfigure[Architecture B (MNIST)]{\includegraphics[width=0.45 \linewidth]{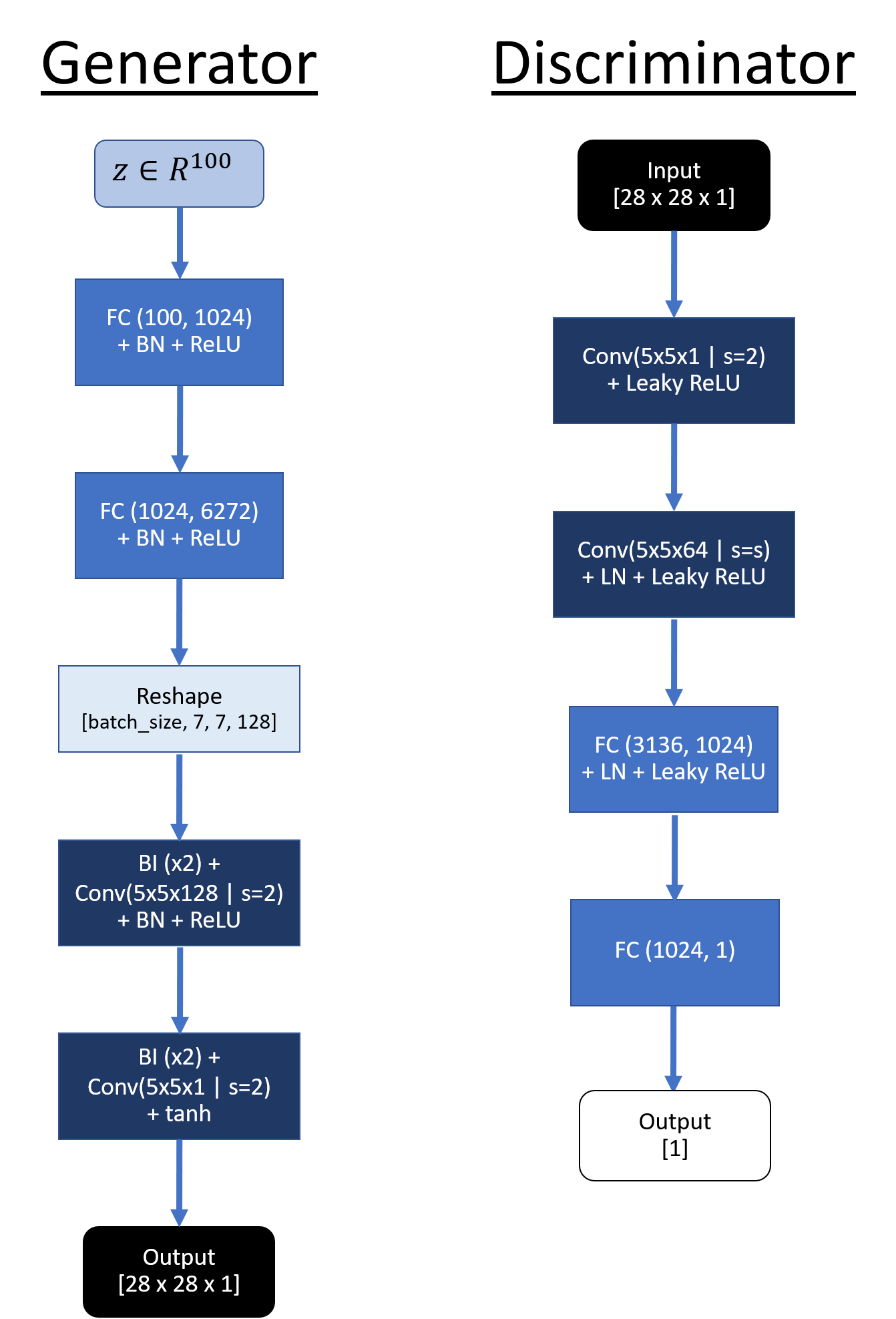}} 
\subfigure[Architecture C (two-phase microstructure)]{\includegraphics[width=0.45 \linewidth]{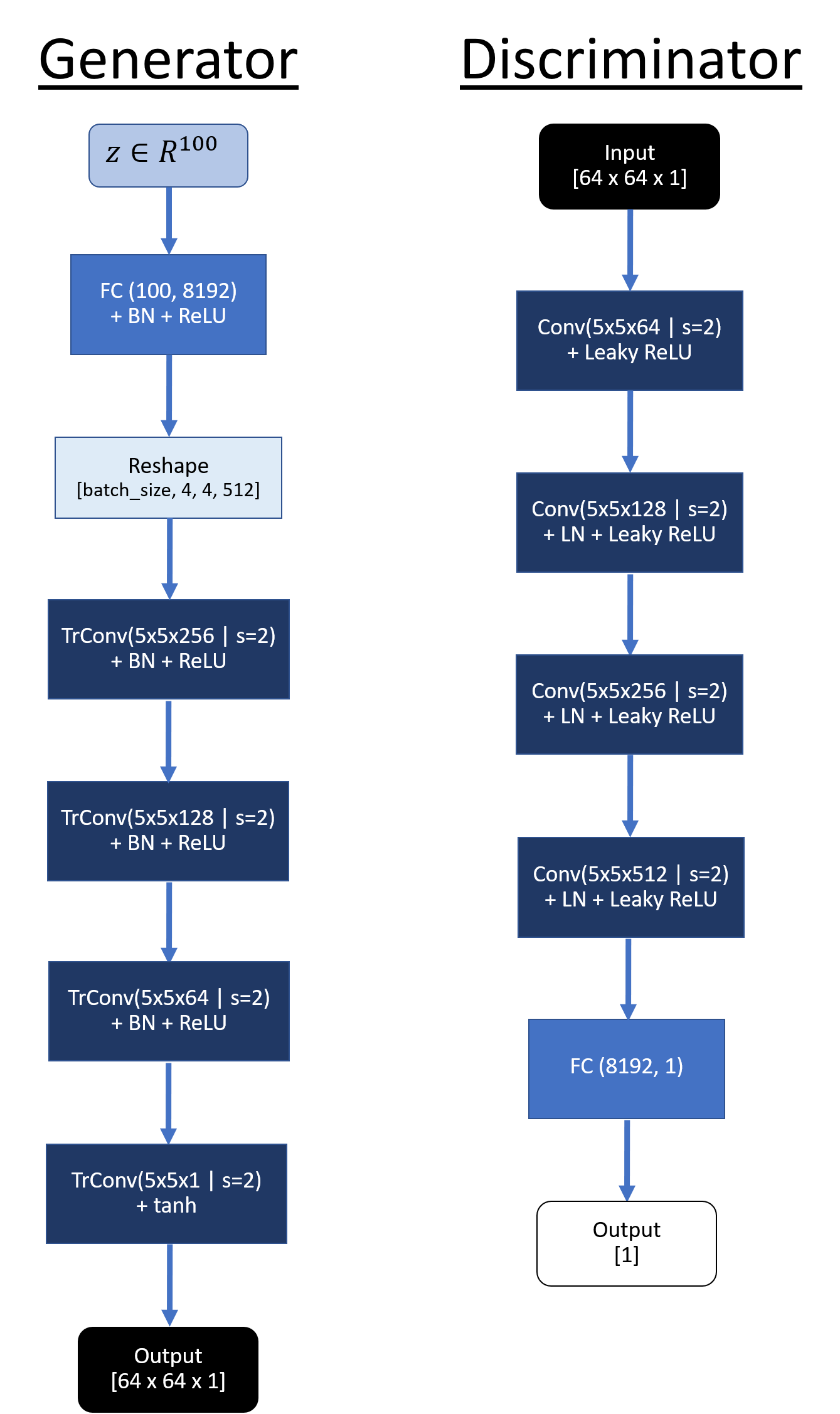}} 
\end{center}
\caption{\label{fig:arch_abc} Generator and discriminator architectures (type A, B, and C) of WGAN-GP for inverse heat conduction experiment.} 
\end{figure}

\begin{figure}[!htbp]
\begin{center}
\subfigure[Architecture D (inverse Radon transform)]{\includegraphics[width=0.46 \linewidth]{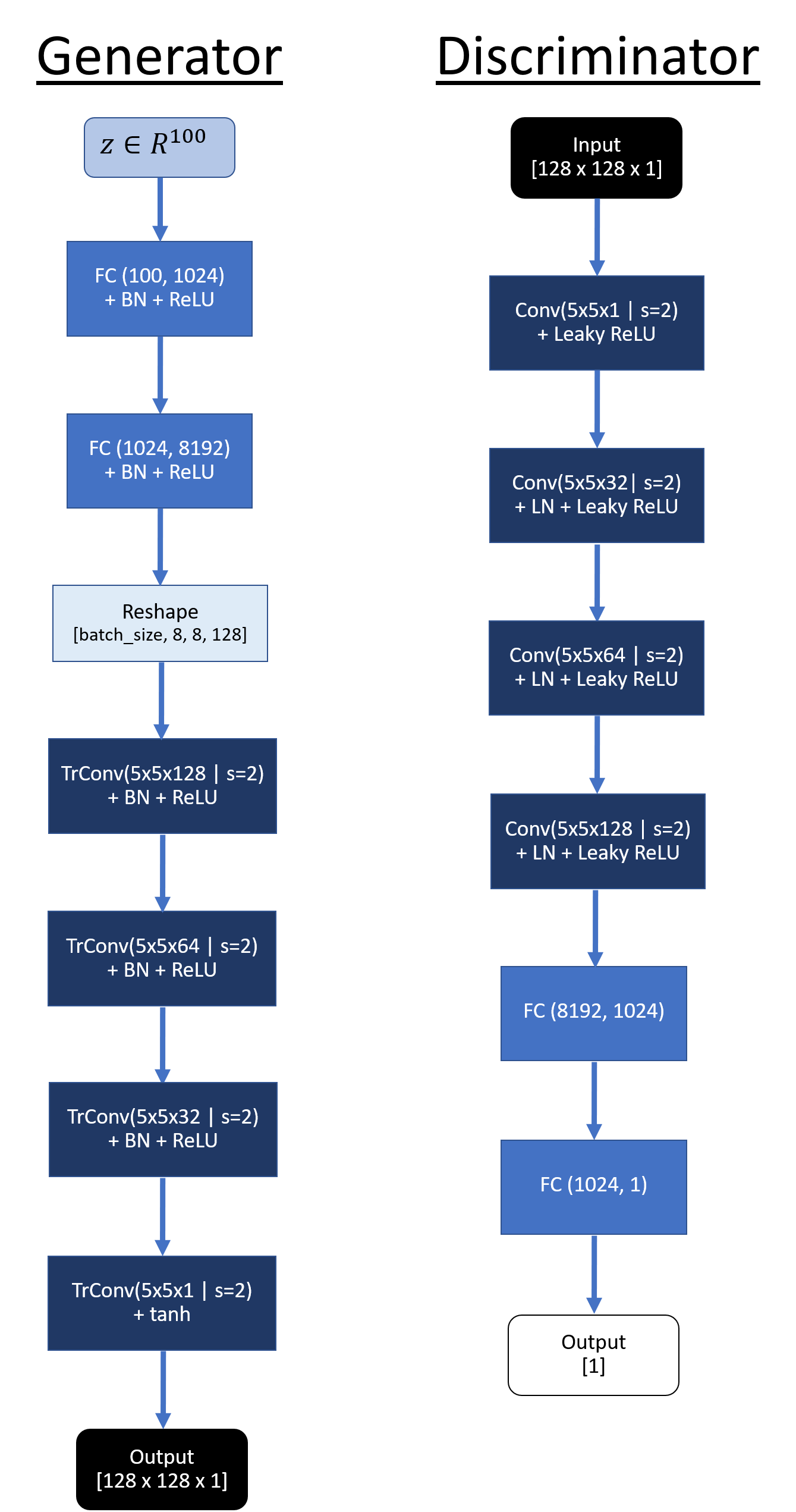}} 
\subfigure[Architecture E (elasticity imaging)]{\includegraphics[width=0.46 \linewidth]{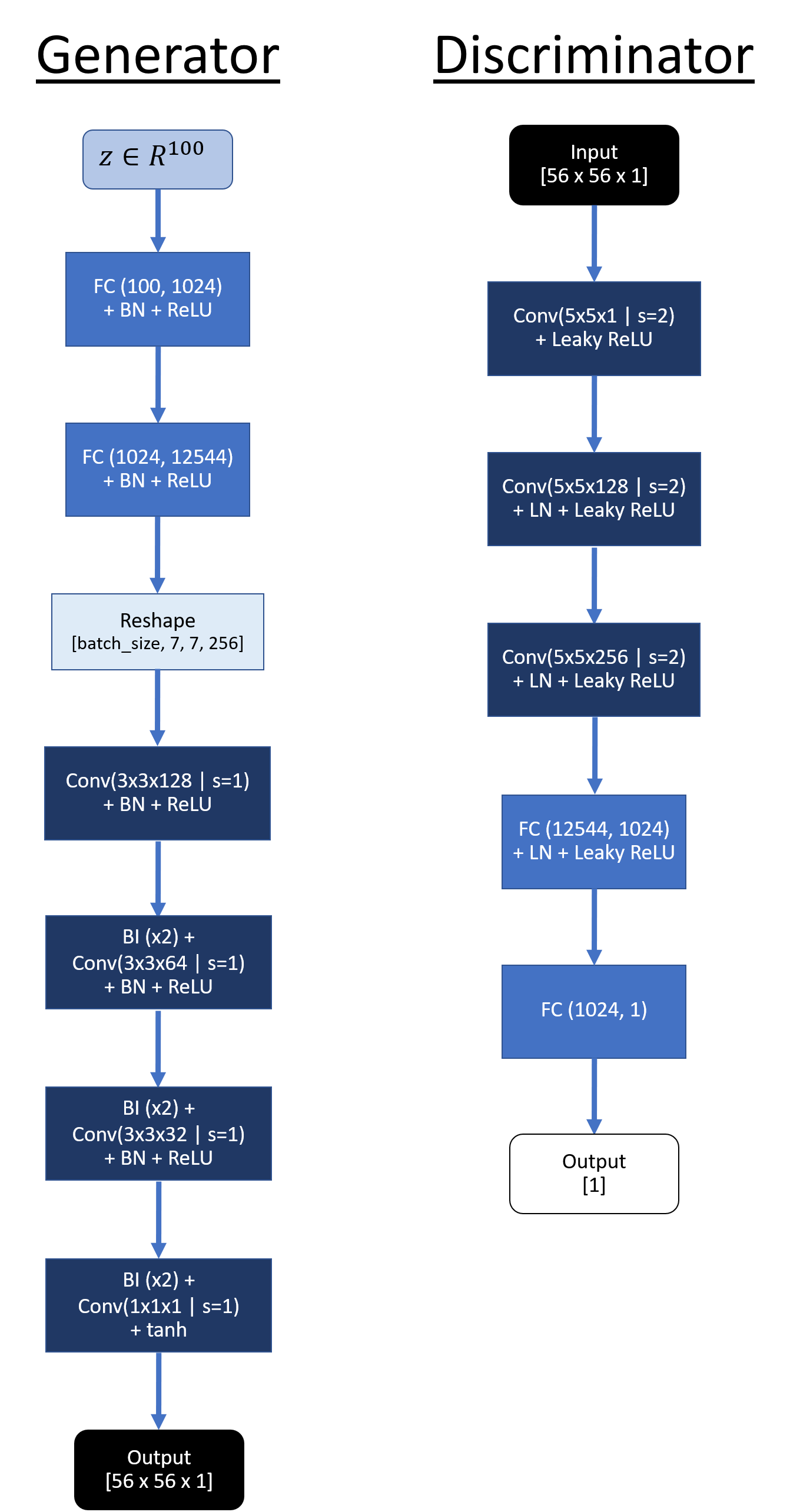}} 
\end{center}
\caption{\label{fig:arch_de}  Generator and discriminator architectures of WGAN-GP for inverse Radon transform and elasticity imaging experiment respectively.} 
\end{figure}

In this section we provide details about different modeling choices and hyper-parameters used in this study. In all numerical experiments we use WGAN-GP \cite{gulrajani2017improved} for learning the prior density. The detailed architecture and associated hyper-parameters used in training of this model for different numerical experiments is provided in Table \ref{tab:hparams} and Figures \ref{fig:arch_abc} and \ref{fig:arch_de}. Once the  prior density is learned, posterior sampling is performed using Hamiltonian Monte Carlo method. The associated hyper-parameter values for this step is also provided in Table \ref{tab:hparams}.

\begin{table}[!htbp]
\renewcommand{\arraystretch}{1.5}
\centering
\caption{Hyper-parameters for WGAN-GP and MCMC}
\begin{adjustbox}{width=\linewidth}
\begin{tabular}{c c c c c c c c}
\toprule
& & Inverse problem & \multicolumn{3}{c}{Inverse Heat Conduction} & Inverse Radon transform & Elastography \\
\cmidrule{4-6}
& &
\multirow{2}{*}{Dataset} &   \multirow{2}{*}{Rectangular} &   \multirow{2}{*}{MNIST} & Cahn-Hilliard & Shepp-Logan & Circular \\
& & & & & microstructure & phantom & phantom \\

\midrule
\parbox[t]{2mm}{\multirow{8}{*}{\rotatebox[origin=c]{90}{Learning prior}}} & & Architecture & Type A & Type B & Type C & Type D & Type E \\
& & Epochs & 100 & 1000 & 500 & 500 & 500 \\
& & Learning rate & 0.0002 & 0.0002 & 0.0001 & 0.0002 & 0.0002\\
& & Batch size & 64 & 64 & 64 & 100 & 64\\
& & $n_{critic}/n_{gen}$ & 2 & 5 & 5  & 4 & 5\\
& & Optimizer & Adam & Adam & Adam & RMSProp & RMSProp \\
& & Optimizer params.  & \multirow{2}{*}{0.5, 0.999} & \multirow{2}{*}{0.5, 0.999} & \multirow{2}{*}{0.5, 0.999} & \multirow{2}{*}{0.9} & \multirow{2}{*}{0.9}  \\
\parbox[t]{2mm}{\multirow{3}{*}{\rotatebox[origin=c]{90}{Posterior}}}
& \parbox[t]{2mm}{\multirow{3}{*}{\rotatebox[origin=c]{90}{sampling}}} & ($\beta_1$, $\beta_2$ or decay) & & & \\
& & No. of MCMC samples & 15k & 15k & 25k & 64K & 15k\\
& & Burn-in period & 0.5 & 0.5 & 0.5 & 0.5 & 0.5\\
\bottomrule
\end{tabular}\label{tab:hparams}
\end{adjustbox}
\end{table}

Some notes regarding nomenclature used in these figures:

\begin{itemize}
    \item Conv (H $\times$ W $\times$ C $\vert$ s=n): convolutional layer with filter size of H $\times$ W and number of filters = C with stride (s)=n.
    \item TrConv (H $\times$ W $\times$ C $\vert$ s=n): transposed convolutional layer with filter size of H $\times$ W and number of filters = C with stride (s)=n.
    \item FC (p, q): fully connected layer with p neurons in input and q neurons in output. 
    \item BN : Batch Norm,  LN : Layer Norm, BI (x2) : Bi-linear Interpolation (upscaling by a factor of 2).
    \item LReLU = Leaky ReLU with $\alpha$=0.2.
\end{itemize}

\section{Computing the TV solution}\label{app:TV}
We compare the MAP estimate obtained from our approach with the optimal solution (denoted by $\infv^*$) obtained by solving a deterministic inverse problem with Total Variation (TV) regularization, which is a gold-standard regularizer for physics-driven inverse problems. Specifically, we compute $\infv^*$ by solving following optimization problem,
\begin{equation}
    \infv^* = \frac{1}{2}\int_{\Omega}(\nmeas - \bm{f}(\infv))^2 \bm{dy} + \frac{\gamma}{2}\int_{\Omega} \sqrt{(\nabla\infv)^2 + \epsilon}\  \bm{dx},
    \label{eqn:tv_opti}
\end{equation}
where $\gamma$ is the regularization parameter, $\epsilon$ is total variation parameter that ensures continuous derivatives at $\infv = \bm{0}$, and $\nabla$ denotes an approximation to the gradient of the spatial field represented by $\infv$. 

\begin{figure}[!htbp]
\begin{center}
\subfigure[$\bm{\eta} = 0.01$]{\includegraphics[width=0.46 \linewidth]{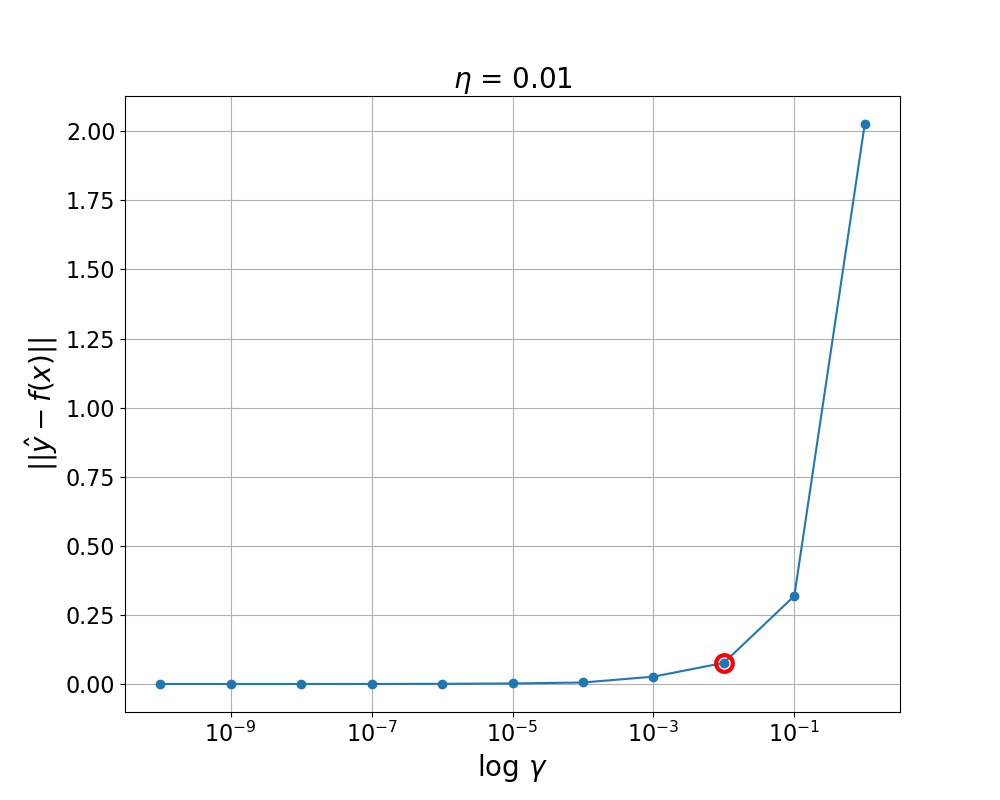}} 
\subfigure[$\bm{\eta} = 0.1$]{\includegraphics[width=0.46 \linewidth]{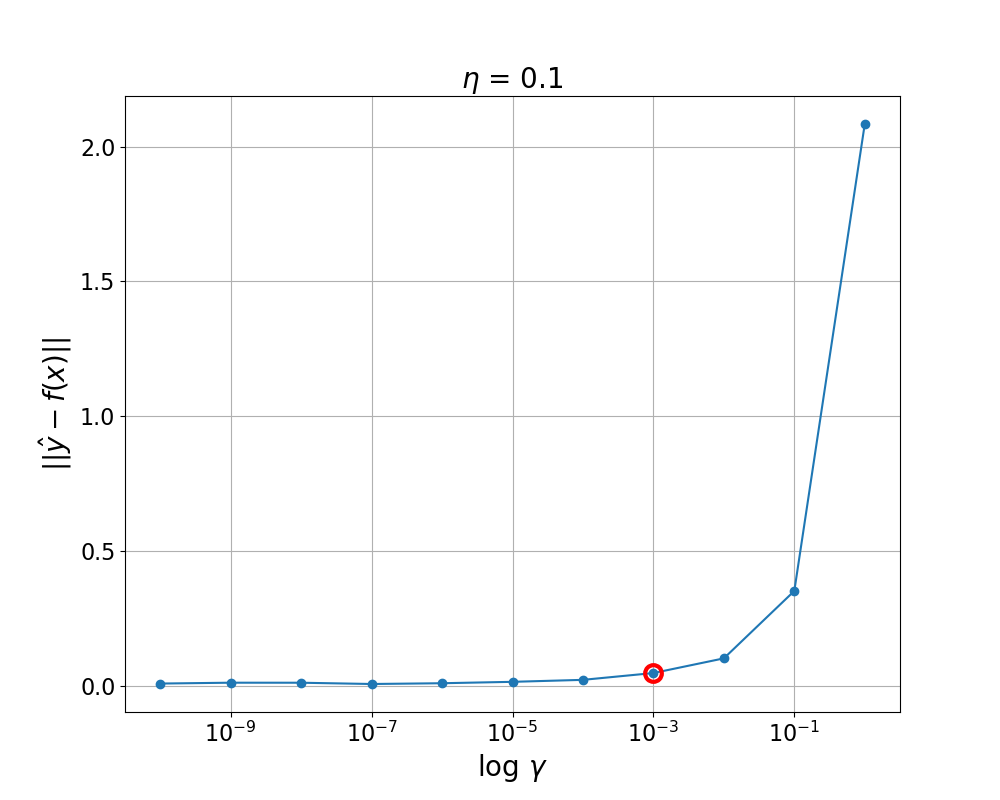}} 
\subfigure[$\bm{\eta} = 1.0$]{\includegraphics[width=0.46 \linewidth]{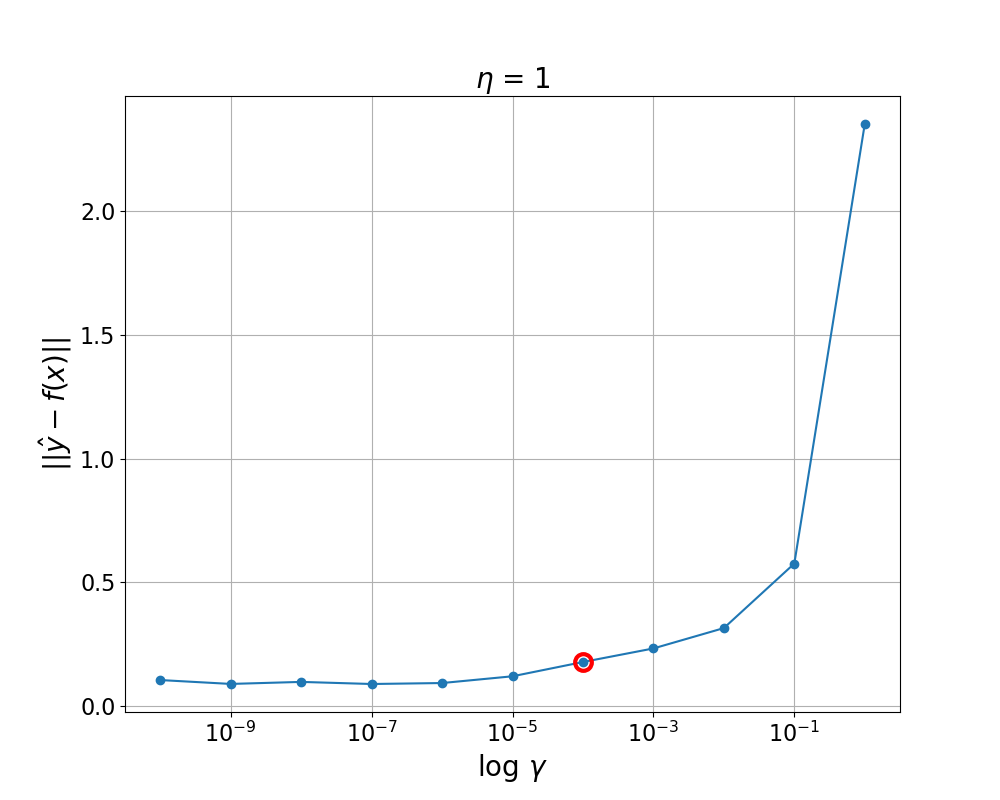}} 
\end{center}
\caption{\label{fig:3Lcurves}  L-curves for   selecting the optimal value of regularization parameter ($\gamma$) for TV prior baseline for the coefficient inversion experiment with MNIST dataset as described in Section \ref{sec:heat_conduction} (Figure \ref{fig:tv_gan_compare}) for different measurement noise levels ($\eta$). Red marker shows the optimal value of $\gamma$ used for producing reconstruction results.} 
\end{figure}

We solve the above optimization problem using inexact Hessian-free Newton-Conjugate Gradient (CG) method following the strategy described in \cite{Petra2012inexact}. The main components of this method are as follows:
\begin{itemize}
    \item The method approximately solves the linear system stemming from each Newton iteration of the optimization problem described in equation \eqref{eqn:tv_opti} using the conjugate gradient method.
    \item The Newton system is solved inexactly by early termination of CG iterations via Eisenstat-Walker and Steihaug criteria.
    \item The method does not require explicit calculation of expensive Hessian matrix for Newton iterations. Instead it requires first solving the forward and adjoint problems to compute the gradient, and then incremental forward and incremental adjoint problems to compute the action of Hessian on gradient.
    \item The step size for the optimization algorithm is selected by Armijo backtracking line search.
    \item The optimal value of the regularization parameter $\gamma$ is selected based on the  L-curve criteria \cite{Calvetti2000} and the value of TV parameter $\epsilon$ is set to $1e-4$. In Figure \ref{fig:3Lcurves} we provide L-curves for the thermal conductivity inversion experiment for the MNIST dataset described in Section \ref{sec:heat_conduction} as a reference. Specifically we provide three L-curves for the experiment described in Figure \ref{fig:tv_gan_compare} for three different measurement noise levels. Based on this L-curve criteria, the optimal value of $\gamma$ (shown with red marker points in Figure \ref{fig:3Lcurves}) is selected near the ``knee'' of this curve and is used for producing TV-based MAP results.
\end{itemize}

\section{Datasets}
In this section we provide details about datasets used in different numerical experiments. We explain details of how different datasets were generated and the values of associated parameters.

\subsection{Original Shepp-Logan phantom}\label{app:phantom}
\begin{figure}[htbp]
\begin{center}
\includegraphics[width=0.50\textwidth]{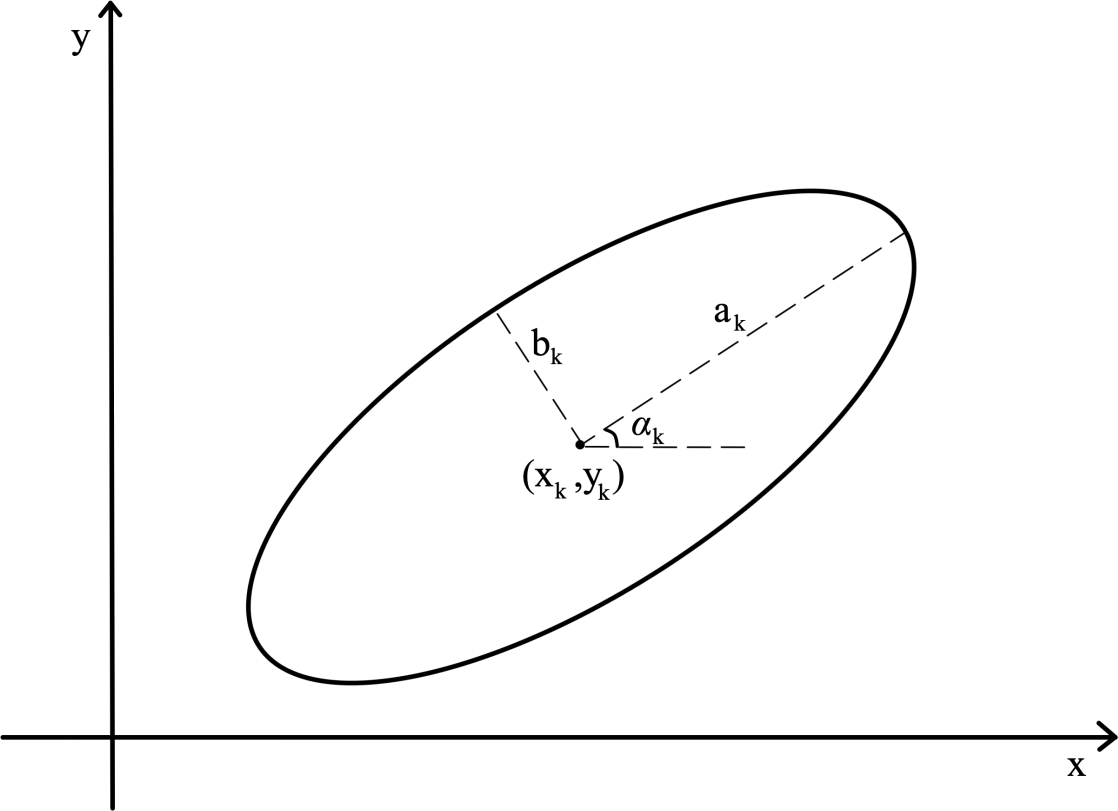}
\caption{Ellipse used to construct the Shepp-Logan phantom.}
\label{fig:ellipse}
\end{center}
\end{figure}

The Shepp-Logan phantom is composed of a union of ten ellipses. The $k$-th ellipse $E_k$ (see Figure \ref{fig:ellipse}) is centered at $(r_k,s_k)$, with semi-axis lengths $a_k$, $b_k$ and angle of inclination $\alpha_k$. The density inside this ellipse is given by a constant $\rho_k>0$, while the density outside the ellipse is 0. The base parameter values of each ellipse is listed in Table \ref{tab:sl_params}. The density of the phantom at any coordinate $(r,s)$ is given by 
\begin{equation}\label{eqn:centered_sl}
\rho(r,s) = \sum_{k=1}^{10} C_k(r,s), \qquad C_k(r,s) = \begin{cases} \rho_k  & \quad \text{if } (r,s) \in E_k\\
0 & \quad \text{otherwise} \end{cases}
\end{equation}

\begin{table}[!htbp]
\centering
\begin{tabular}{|l|l|l|l|l|l|l|}
\hline
$k$ & $r_k$ & $s_k$   & $a_k$  & $b_k$ & $\alpha_k$ (degrees) & $\rho_k$ \\ \hline
1   & 0.0   & 0.0     & 0.69   & 0.92  & 0                    & 1.0      \\ \hline
2   & 0.0   & -0.0184 & 0.6624 & 0.874 & 0                    & -0.8     \\ \hline
3   & 0.22  & 0.0     & 0.11   & 0.31  & -18                  & -0.2     \\ \hline
4   & -0.22 & 0.0     & 0.16   & 0.41  & -18                  & -0.2     \\ \hline
5   & 0.0   & 0.35    & 0.21   & 0.25  & 0                    & 0.1      \\ \hline
6   & 0.0   & 0.1     & 0.046  & 0.026 & 0                    & 0.1      \\ \hline
7   & 0.0   & -0.1    & 0.046  & 0.046 & 0                    & 0.1      \\ \hline
8   & -0.08 & -0.605  & 0.046  & 0.023 & 0                    & 0.1      \\ \hline
9   & 0.0   & -0.606  & 0.023  & 0.023 & 0                    & 0.1      \\ \hline
10  & 0.06  & -0.605  & 0.023  & 0.046 & 0                    & 0.1      \\ \hline
\end{tabular}
\caption{Base parameters for Shepp-Logan head phantom.}
\label{tab:sl_params}
\end{table}

\subsection{Circular phantom dataset}\label{app:elsto_phantom}
For the elastography experiment described in Section \ref{sec:elastography} we use a circular phantom dataset. This dataset (of shear modulus field) was synthetically generated by varying three parameters: x and y-coordinate of the center of the circular inclusion and the ratio of the shear modulus value inside the inclusion to shear modulus value outside the inclusion. The total domain size is 34.608 $mm$ in x-direction and 26.287 $mm$ in y-direction. x-coordinate and y-coordinates were sampled from $\mathcal{U}[7.30, 19.96]$ $mm$ and $\mathcal{U}[7.4, 25.96]$ $mm$ respectively. The ratio of the shear modulus (value inside the inclusion to background) was sampled from $\mathcal{U}[1, 8]$ with the background value of shear modulus set to 4.7 $kPa$. The choice of the shape of the inclusion and the values of three parameters were made based on the prior knowledge about the experiment. This simulates many real-world scenarios where, based on prior domain knowledge, one selects the appropriate \textit{prior} distribution.

\newpage

 \bibliographystyle{elsarticle-num} 
 \bibliography{cas-refs}





\end{document}